%% file: main.tex
\pdfoutput=1

\documentclass[11pt]{article}

\usepackage{ACL2023}
\usepackage{booktabs}
\usepackage{multirow}
\usepackage{longtable}
\usepackage{ragged2e}  
\usepackage{array}
\usepackage{pifont}
\usepackage{xcolor}
\usepackage{ulem}
\usepackage{cleveref}
\usepackage{lipsum}
\usepackage{subcaption}
\usepackage{times} 
\usepackage{latexsym}
\usepackage{float}
\usepackage{calc}
\usepackage{cuted}      
\usepackage{caption}   
\usepackage{colortbl}
\usepackage[T1]{fontenc}
\definecolor{systemcolor}{rgb}{0.2, 0.7, 0.6} 
\definecolor{gray}{rgb}{0.5,0.5,0.5}
\usepackage{etoolbox}

\usepackage[utf8]{inputenc}

\usepackage{microtype}

\usepackage{inconsolata}

\newcommand{\deemph}[1]{{\color{black!40}#1}}
\crefname{appendix}{App.}{Apps.} 
\Crefname{appendix}{App.}{Apps.} 
\usepackage{graphicx}
\usepackage{longtable, booktabs, array, xcolor}
\graphicspath{ {figures/} }
\graphicspath{ {tables/} }
\usepackage[ruled,vlined,linesnumbered]{algorithm2e}
\SetNlSty{}{}{:}
\DontPrintSemicolon 
\SetAlgoNlRelativeSize{-1}
\SetKwProg{Fn}{def}{\string:}{}
\SetKw{KwNot}{not}
\SetKw{KwNotIn}{not in}
\SetKwFunction{schemaSupervisor}{schema\_supervisor}
\SetKwFunction{parserSupervisor}{parser\_supervisor}
\SetKwFunction{dialogueManager}{dm\_update\_transcript}
\SetKwFunction{executeActions}{execution\_engine}
\SetKwFunction{assertSymbolsInSchema}{assert\_symb\_in\_schema}
\SetKwFunction{pytod}{pytod}
\SetKwFunction{generateSSPrompt}{ss\_prompt}
\SetKwFunction{getPSPrompt}{ps\_prompt}
\SetKwFunction{LLM}{LLM}
\SetKwFunction{assertUserProvidedRequestedSlots}{assert\_slots\_assigned}
\SetKwData{SessionTranscript}{SessionTranscript}
\SetKwData{SystemActions}{SystemActions}
\SetKwData{ProgramStatements}{ProgramStatements}
\SetKwData{AssistantSchema}{Schema}
\SetKwData{SlotList}{SlotList}
\SetKwData{ValueList}{ValueList}
\SetKw{KwIn}{in}
\SetKwBlock{Try}{\textbf{try}}{}
\SetKwBlock{Catch}{\textbf{except}}{}

\definecolor{cambridgeblue}{RGB}{35, 146, 144} 
\definecolor{mustard}{RGB}{254, 174, 0}
\definecolor{customorange}{RGB}{255, 179, 102}
\definecolor{custompurple}{RGB}{127, 0, 255}
\definecolor{customgreen}{RGB}{0, 135, 0}
\definecolor{customred}{RGB}{200, 0, 27}
\definecolor{greenssprompt}{RGB}{0, 138, 0}
\newcommand{\greenuline}[1]{\textcolor{greenssprompt}{\uline{#1}}}

\title{PyTOD: Programmable Task-Oriented Dialogue with Execution Feedback}

\author{
    \textbf{Alexandru Coca\textsuperscript{1}}{\thanks{\; Work done while at Apple.}},
    \textbf{Bo-Hsiang Tseng\textsuperscript{2}},
    \textbf{Pete Boothroyd\textsuperscript{2}},
    \textbf{Jianpeng Cheng\textsuperscript{3*}},
    \textbf{Mark Gaynor\textsuperscript{2}},\\
    \textbf{Zhenxing Zhang\textsuperscript{2}},
    \textbf{Joe Stacey\textsuperscript{4*}},
    \textbf{Tristan Guigue\textsuperscript{2}},
    \textbf{Héctor Martinez Alonso\textsuperscript{2}}, \\
    \textbf{Diarmuid Ó Séaghdha\textsuperscript{2}}, 
    \textbf{Anders Johannsen\textsuperscript{2}}
    \\
    \\
    \textsuperscript{1}University of Cambridge,
    \textsuperscript{2}Apple,                                          
    \textsuperscript{3}Meta,
    \textsuperscript{4}Imperial College London
    \\ 
    \texttt{ac2123@cam.ac.uk}, \texttt{ajohannsen@apple.com}
}

\begin{document}
\maketitle
\input{sections/0_abstract/v0.1}
\input{sections/1_introduction/v0.4.tex}
\input{sections/simulation/simulation.tex}

\input{sections/method/v0.2}
\input{sections/experiments/v0.1}

\input{sections/results/v0.1}
\input{sections/analysis/v0.1}
\input{sections/related_work/v0.1}

\section{Conclusion}
We introduced PyTOD, a model that generates code incrementally, tracks state through execution, and uses policy and execution feedback to automatically generate error-correcting prompts. By coupling state tracking with execution, PyTOD shows enhanced cross-turn consistency and thus improved real-world reliability. We release \texttt{pytodlib}, a simulation grounding SGD, to advance research on zero-shot TOD agents and conversational tool use. Future work will focus on enhancing PyTOD's robustness to copy errors and applying it to LLM-based zero- and few-shot E2E dialogue modeling. 
\section*{Limitations}

\paragraph{API retrieval} Like most state-of-the-art approaches on the SGD and MultiWOZ datasets, PyTOD assumes knowledge of the service schema at each turn (but not the API). In real-world scenarios, however, virtual assistants must first infer the schema before tracking dialogue state. This is particularly challenging in SGD, where multiple services within the same domain exhibit fine-grained differences. For instance, \texttt{Buses\_1} and \texttt{Buses\_2} both implement the \texttt{FindBus} APIs, yet disambiguation between the two services is only possible if users mention \texttt{fare\_type}, an \textit{optional} slot. Similarly, the \texttt{Buses\_3} service (from the test set) can only be distinguished if the \texttt{category} slot is provided. As discussed in \S \ref{sec:err-analysis}, PyTOD relies on accurate intent parsing, making it susceptible to service disambiguation errors that will degrade performance.

\paragraph{Schema robustness} We have not explicitly evaluated PyTOD's robustness to linguistic schema variations, which are known to affect transfer learning-based DST systems. While AP accuracy will degrade, PyTOD's SS - the primary contributor to its performance (\S \ref{sec:ablation}) - is expected to mitigate this impact since it performs zero-shot corrections using MQA prompts. Future work could enhance AP and PS robustness using transfer learning from QA task \cite{qa-cross-domain-lin, cho-etal-2023-continual}, knowledge-seeking turn grounding \cite{mypaper} or synthetic schemas \cite{coca-etal-2023-robust}, none of which require additional annotation. 

\paragraph{Prompt optimisation} While PS and AP latency are optimized through dialogue history truncation and minimal generation length, the AP prompt itself remains unoptimised. The AP header already maintains a stack of completed tasks and retrieved entities, effectively summarising the dialogue history. However, our prompts contain both the header and transcript, introducing redundancy and increasing system latency. 

Additionally, the API header presents an opportunity for personalised conversational intelligence. Embedding and storing it in a vector database could enable retrieval across multi-session conversations, improving continuity and personalisation. We will explore this in future work.

\paragraph{Grammar-constrained decoding} We opted for a deep learning approach to constraining AP output to the schema due to the complexity of working with dynamically generated grammar rules needed to constrain decoding to a set of valid tokens. We considered this advanced optimisation a research topic for future work.

\paragraph{Interactive evaluation}Our results demonstrate that state-of-the-art DST models exhibit stability errors when evaluated using C-JGA. As DST systems improve, we believe evaluating models in real-world user interactions will be essential for assessing their practical viability.

\subsection*{Acknowledgments}

Alexandru Coca's doctoral studies are supported by the EPSRC grant EP/R513180/1. He would like to acknowledge generous financial support from the Cambridge Philosophical Society through the Research Studentship. The authors thank Ankit Bhattarai from the University of Cambridge for his contribution to the SDT-Seq replication study and evaluation.  Xiaofeng Wu and Stephen Pulman from Apple alongside anonymous reviewers are acknowledged for their improvement suggestions.
\bibliography{anthology,custom}
\bibliographystyle{acl_natbib}

\appendix
\input{appendices/simulation}
\input{appendices/prompt_details_v0.3}
\input{appendices/experimental_setup_v0.2}

\input{appendices/additional_results_v0.1}

\input{appendices/annotated_transcripts}

\end{document}

%% file: sections/0_abstract/v0.1.tex
\begin{abstract}

Programmable task-oriented dialogue (TOD) agents enable language models to follow structured dialogue policies, but their effectiveness hinges on accurate state tracking. We present PyTOD, an agent that generates executable code to track dialogue state and uses policy and execution feedback for efficient error correction. To this end, PyTOD employs a simple constrained decoding approach, using a language model instead of grammar rules to follow API schemata. This leads to state-of-the-art state tracking performance on the challenging SGD benchmark. Our experiments show that PyTOD surpasses strong baselines in both accuracy and robust user goal estimation as the dialogue progresses, demonstrating the effectiveness of execution-aware state tracking.
\end{abstract}

%% file: sections/1_introduction/v0.4.tex
\section{Introduction}
\label{sec:intro}
TOD agents provide natural language interfaces which enable users to control their digital environment to complete daily tasks. Such agents typically include a \textit{dialogue state tracking} (DST) component, which maps the conversation history to a symbolic representation of the task-relevant information communicated during the exchange. At each turn, a \textit{dialogue manager} (DM) uses this information to take \textit{system actions}\footnote{For example, retrieving information from a knowledge base or prompting the user to provide task constraints.} necessary to help the user complete the task. The agent behaviour is controlled by a \textit{dialogue policy} defined by application developers. 
\input{figures/pytod_overview}
 
Domain adaptation has long been a challenge for state tracking \cite{jacqmin-2022-est}, dialogue management \cite{stardataset} and end-to-end (E2E) \cite{AnyTOD} agents, as it often requires developers to collect and annotate new datasets for retraining. To address this challenge, fine-tuning pre-trained language models (PLMs) \cite{GPT2, T5} within the schema-guided paradigm \cite{SGD, stardataset} has emerged as a powerful approach. Schemata define the APIs accessible to the agent, including textual descriptions of their functions and parameters (referred to as \textit{slots}). \citet{stardataset} extend schemata by including descriptions of the actions that TOD agents can perform. \citet{AnyTOD} build on these advances with AnyTOD, a state-of-the-art (SOTA) neuro-symbolic agent capable of following dialogue policies unseen during fine-tuning.

AnyTOD uses the schema and dialogue history to first generate a symbolic \textit{state sequence}. This sequence identifies the API the user wishes to interact with and the slot values the user has mentioned. A second symbolic sequence - encoding the actions the user\footnote{For example, providing a slot value or requesting information about a knowledge base item retrieved by the system.} and system have taken - is generated sequentially after the state sequence. Both sequences are subsequently interpreted by a \textit{deterministic policy program} which recommends the next system action. State sequence is thus critical: prediction errors can prevent the system from taking the correct actions, leading to breakdowns in the interaction. 

Despite its strengths, AnyTOD has some limitations. First, it re-estimates the state and action history at every dialogue turn solely based on the dialogue history and schema, which increases generation length and amplifies the risk of state-tracking errors. Second, it fails to exploit previous system actions to verify the correctness of the state sequence. Finally, its reliance on symbolic representations of state and action sequences requires additional system components to translate them into executable code, adding deployment complexity. We present PyTOD, a programmable dialogue system that addresses these challenges by directly communicating with its execution environment and policy programs to perform accurate state tracking for unseen APIs and domains, as discussed next.

\textbf{PyTOD overview} PyTOD incrementally generates dialogue states as code, using policy- and execution feedback for accurate state estimation (\Cref{fig:pytod}). It operates as follows: (1) an action parser (AP) (\S \ref{sec:ap}) processes the user query to produce one or more Python statements; (2) the dialogue manager (DM) (\S \ref{sec:dm}) executes these statements, ensuring they valid according to the schema; (3) if invalid slots are found, the DM calls the schema supervisor (SS) (\S \ref{sec:ss}), a language model that constrains code generation according to the decoding schema; and (4) the DM evaluates the constrained output by comparing expected and current environment states, invoking a parser supervisor (PS) (\S \ref{sec:ps}) to recover from possible omissions or semantic errors. The dialogue state is derived by executing the program generated by PyTOD at each user turn. 
 
\textbf{Contributions} Unlike PyTOD, TOD agents often optimise DST and DM independently, overlooking how policy information can enhance state generation accuracy. We further demonstrate that feedback from the execution environment enables language models to constrain decoding without requiring additional training data, with minimal developer effort and only a slight increase in system latency. PyTOD achieves SOTA performance on the challenging Schema-Guided Dialog (SGD) dataset \cite{SGD}.

As an additional contribution, we release \texttt{pytodlib}\footnote{Code: \url{https://github.com/apple/ml-pytod}.} , a Python library that simulates SGD APIs, including database responses and API behavior. This toolkit addresses the scarcity of conversational tool-use corpora, providing a valuable resource for benchmarking large language models' ability to handle complex, multi-task, goal-oriented conversations while following predefined policies. 

%% file: figures/pytod_overview.tex
\begin{figure}[h!]
\vspace{-0.3cm}
\includegraphics[width=0.88\linewidth, trim={8mm 9mm 7mm 9mm}, clip]{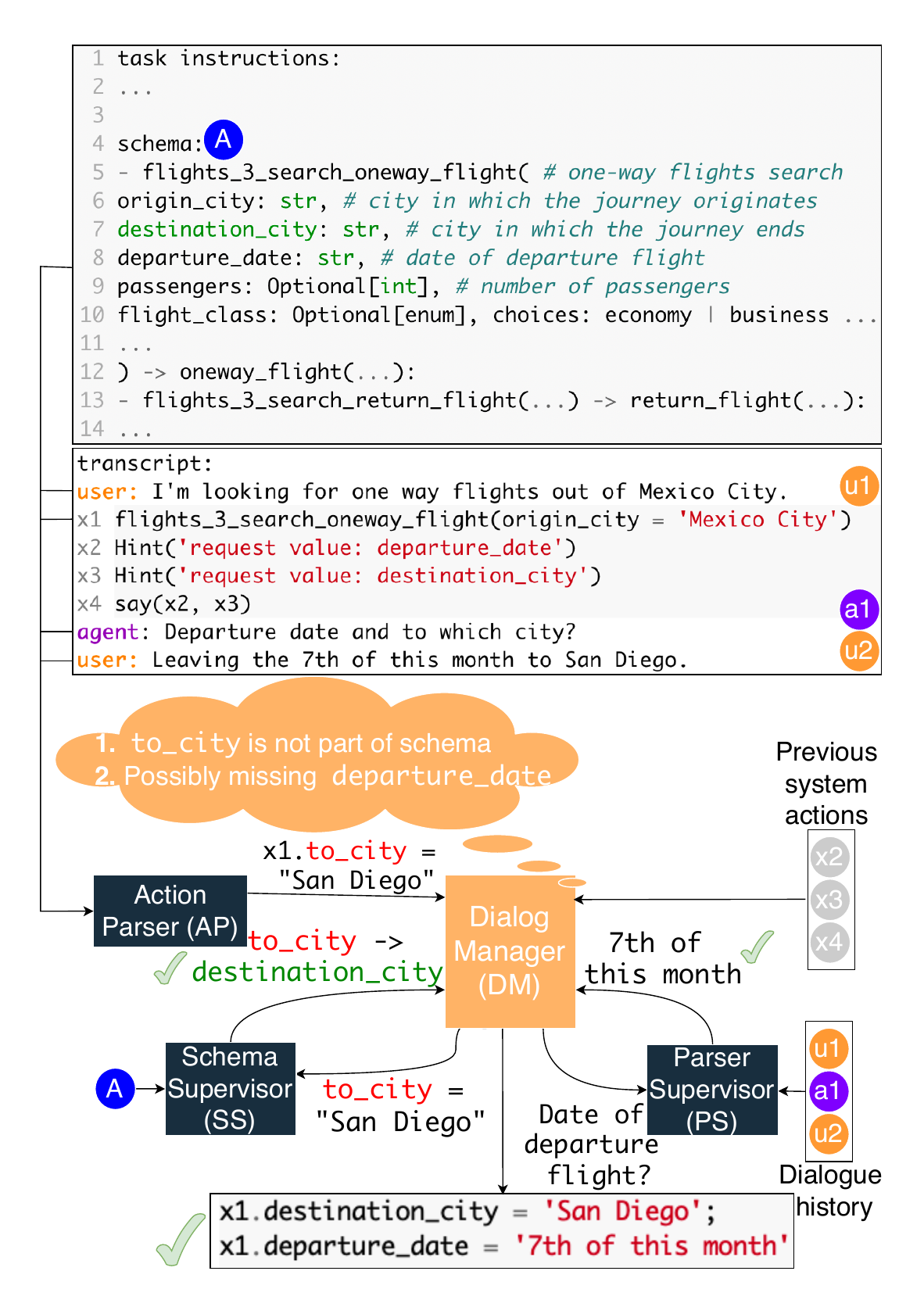}
    \centering
    \caption{PyTOD overview. The action parser generates Python statements ({x1.\textcolor{red}{to\_city} = "San Diego"}) representing the actions the user took at the current turn (\textcolor{customorange}{u2}) given API schemata (\textcolor{blue}{A}), dialogue history (\textcolor{customorange}{u1},  \textcolor{custompurple}{a1}) and previous user actions (\deemph{x1}). The dialogue manager (DM) executes the user action in a simulated environment. A schema supervisor is invoked by the DM to correct errors if predicted statements contain slot names that are not part of the API schema (e.g., the slot \textcolor{red}{to\_city} is mapped to \textcolor{customgreen}{destination\_city}, a member of the schema).  Given knowledge of previous system actions (\deemph{x2}, \deemph{x3}) the DM detects slot omissions and invokes a parser supervisor to correct them (e.g., flight date is recovered).}
    \label{fig:pytod}
    \vspace{-0.55cm}
\end{figure}

%% file: sections/simulation/simulation.tex
\section{Simulated Environment}

As a programmable dialogue system, PyTOD relies on a simulation of its operating environment and an execution engine, described in this section. Our simulation is based on SGD due to its large ontology, complex policy and high-quality annotations.

\subsection{API simulation}
\label{sec:api-sim}
We simulate the $58$ APIs from the validation and test splits. Each API is implemented as a Python object that can be instantiated and updated by a dialogue agent.

Our simulation is based on a detailed analysis of the system and user dialogue acts in the SGD dataset. Accordingly, upon instantiation or update, the simulation returns structured \textit{system action recommendations} or \textit{hints}, that the agent can execute to continue the dialogue. 

The specific action recommendations depend on the dialogue state. If the user has not specified all the required arguments for API execution, \textit{slot filling hints} (x2, x3 in \Cref{fig:pytod}) guide the agent towards requesting additional information from the user. When all task details are known, search queries return \textit{entities} retrieved from databases, whereas for transactions, \textit{confirmation hints} (x17 in \Cref{fig:transcript}, turn 7) are suggested to confirm slot values.  Upon execution, the APIs may return \textit{notifications} if the task could not be completed as requested by the user and, in some cases, \textit{alternative hints} to guide the agent to propose satisfiable task constraints (\Cref{fig:transcript}, turn 8). Finally, agent initiative is simulated via related-task suggestions (\Cref{fig:transcript}, turn 3) and \textit{user prompt hints} (\Cref{fig:transcript}, turn 7). See \Cref{appendix:apis-sim} for implementation details.

External API and database responses are simulated using the semantic and entity annotations. The slot values extracted from the dialogue context are normalised prior to invoking APIs, allowing the simulation to return entities as Python objects that dialogue agents can inspect and reason over.

\subsection{Execution engine} 
\label{sec:exec-engine}
Code execution proceeds as follows:
\begin{enumerate}
    \item A generated string is parsed to an Abstract Syntax Tree (AST) using Python's \texttt{ast} library
    \item The AST is converted into a Python callable
    \item The callable is invoked, updating the state of the  referenced APIs.
\end{enumerate}
See \Cref{appendix:execution_engine} for further details and examples.

%% file: sections/method/v0.2.tex
\section{PyTOD}
\label{sec:method}
\subsection{Action parser}
\label{sec:ap}
The AP parses user utterances into Python expressions, which are executed to carry out the user's actions. The prompt consists of: (1) a header containing task-specific instructions and linearised schema, alongside a list of completed tasks and entities (e.g., flights) returned by them (\S \ref{sec:header}); (2) a session transcript, where user and system turns are interleaved with Python code snippets representing user actions and execution outputs (\S \ref{sec:transcript}); and (3) context-dependent instructions, dynamically rendered to provide additional task guidance and entity definitions as the conversation progresses (\S \ref{sec:dynamic-instr}).
\subsubsection{Header}
\label{sec:header}
At the start of the conversation, the header instructions (\Cref{fig:pytod}, top) prompt the PLM to identify the API that aligns with the user’s intent and extract any arguments specified by the user (\Cref{fig:task_instruction}, \Cref{appendix:header_prompt}). Schema API definitions are presented next, linearised as Python function signatures (\Cref{fig:pytod}, lines 4–14). Each API name is followed by an \textit{intent description} summarising its function (line 5), while arguments are annotated with their types and descriptions (lines 6–10). For \textit{categorical slots}, which take \textit{closed values} from a predefined set, the argument descriptions are prefixed with the list of valid options (line 10). Return types specify the entities produced by APIs, excluding their properties (line 12), which are displayed dynamically as entities are returned (\S \ref{sec:dynamic-instr}).

As tasks are completed, the header is updated with descriptions of completed tasks and relevant entity definitions (\Cref{fig:task_stack}, \Cref{appendix:header_prompt}).
\input{figures/transcript_representation_v0.6}

\subsubsection{Session transcript}
\label{sec:transcript}
The session transcript records the dialogue history, interleaved with \textit{PyTOD program statements} that capture user and system actions (\Cref{fig:transcript}). 

\input{figures/user_actions}

PyTOD program lines start with \textit{intermediate variables} \cite{LUCID} to enable incremental updates to the dialogue state based on user actions. This design minimises generation length and eschews re-estimating the entire dialogue state after every turn. Moreover, unlike SOTA DST approaches which extract slot values from both agent and user utterances, agent-mentioned slot values are extracted from the database results and API responses so are guaranteed to be correct. These design choices mitigate the impact of copy errors, which typically affect the accuracy of generative DST models \cite{cho-etal-2022-know}.

\textbf{User actions} The user actions PyTOD can parse are listed \Cref{tab:tod_syntax_final}, with syntax examples in \Cref{fig:transcript}. While the actions are closely related to the dialogue acts commonly defined in TOD corpora, \Cref{appendix:session_transcript_prompt} shows the proposed sub-language can represent more complex task-oriented interactions.

\textbf{Communicative system actions} Executing user actions updates the dialogue state. Furthermore, as discussed in \S \ref{sec:api-sim}, the dialogue policy simulation returns system action recommendations based on the dialogue state. The agent selects appropriate actions by passing them as positional arguments to the \texttt{say} routine (e.g., turns 1 and 3 in \Cref{fig:transcript}). \texttt{say} is an interface to the NLG module, which generates the next agent utterance given the selected action recommendations. For efficiency, the \texttt{say} calls and action recommendations are suppressed from the transcript after agent utterance generation -- only the agent utterances are included in the session transcript history.

\textbf{Non-communicative system actions} are persistent in session history and provide cues for state tracking and language generation. These include database call markers which signal entity retrieval and guide the model to predict subsequent iteration calls (x2 in \Cref{fig:transcript}) and the \texttt{perform} statement (x32 in \Cref{fig:transcript_details}, \Cref{appendix:header_prompt}) which marks successful execution of transactional API calls.

We provide further session transcripts with detailed explanations of PyTOD operation in \Cref{appendix:annotated-session-transcripts}.

\subsubsection{Context-dependent instructions}
\label{sec:dynamic-instr}
Context-dependent instructions are rendered dynamically based on generated statements. They list properties of retrieved entities which allow PyTOD to answer questions about them (e.g., turn 3 \Cref{fig:transcript}; see \Cref{fig:cd_instr_qa} in \Cref{appendix:ap-prompt} for prompt) and may encode policy (see \Cref{fig:cd_instr_policy}, \Cref{appendix:ap-prompt}).

\subsection{Dialogue manager}
\label{sec:dm}
The DM coordinates interaction with the user by taking actions recommended by the dialogue policy upon AP output execution. To ensure successful execution, it constrains and validates the AP outputs as described in the reminder of this section.
\subsubsection{AP output constraints}
AP-generated statements must be valid Python expressions with declared variables to execute. The DM enforces these constraints, inserting \texttt{parse\_error}s into the transcript if parsing fails.  Additionally, it restricts generated API names to those listed in the AP header, correcting errors by minimizing edit distance. Once constrained, statements are executed. If the AP predicts a slot name outside the decoding schema, the DM invokes its SS component to constrain it accordingly.

\subsubsection{Schema supervisor}
\label{sec:ss}

The SS generates a prompt using the schema and the AP output, based on three generic templates (\Cref{appendix:ss-templates}). It follows a multiple-choice question answering (MQA) format (\Cref{fig:ss_prompt}), where the question and options depend on the AP error.  If the AP predicts an unknown slot, the prompt lists the all the slot names in the schema with descriptions\footnote{Descriptions are replaced with data type for integer-valued slots and value enumeration for categorical slots.} as answer choices, instructing the model to select the option corresponding to the slot which best matches the AP output (\Cref{fig:unk_slot_prompt}). If the slot name is unknown but its value is listed in the schema, the prompt includes only categorical slot definitions and possible values (\Cref{fig:unk_slot_cat_value_prompt}). For cases where the AP outputs a slot from a training schema but not the current task schema, the model is presented with slot descriptions from the task schema and instructed to select the closest paraphrase (\Cref{fig:memorisation_prompt}).
\input{figures/ss_prompts_v0.2}
\input{figures/ps_prompt}

\subsubsection{Parser supervisor}
\label{sec:ps}
The DM invokes the PS when SS-constrained AP assignment expressions fail to provide values for slots the system previously requested from the user. In response, the PS generates a prompt using the schema and dialogue history, based on a simple template (\Cref{appendix:ps-template}). The prompt follows an extractive QA format (\Cref{fig:ps_prompt}), where the questions correspond to the schema descriptions of the omitted slots (e.g., \textit{Departure date of the flight?} in \Cref{fig:pytod}); the context for answering them is limited to the dialogue history of the current task.

The PS corrects slot omissions and semantic errors. First, it extracts values for all requested slots missing from the AP output. Then, for open-value slots, if a predicted answer exactly matches a value already assigned to a slot in the constrained AP output, the system assumes a semantic error and replaces the predicted slot name with the corresponding omitted one. If no such match is found, the PS appends assignment expressions that bind the predicted answers to the omitted slots and updates the current task (viz \textit{departure\_date}, \Cref{fig:pytod}). 

%% file: figures/transcript_representation_v0.6.tex
\begin{figure}[t]
\includegraphics[width=0.95\linewidth, trim={23mm 0mm 23mm 0mm}, clip]{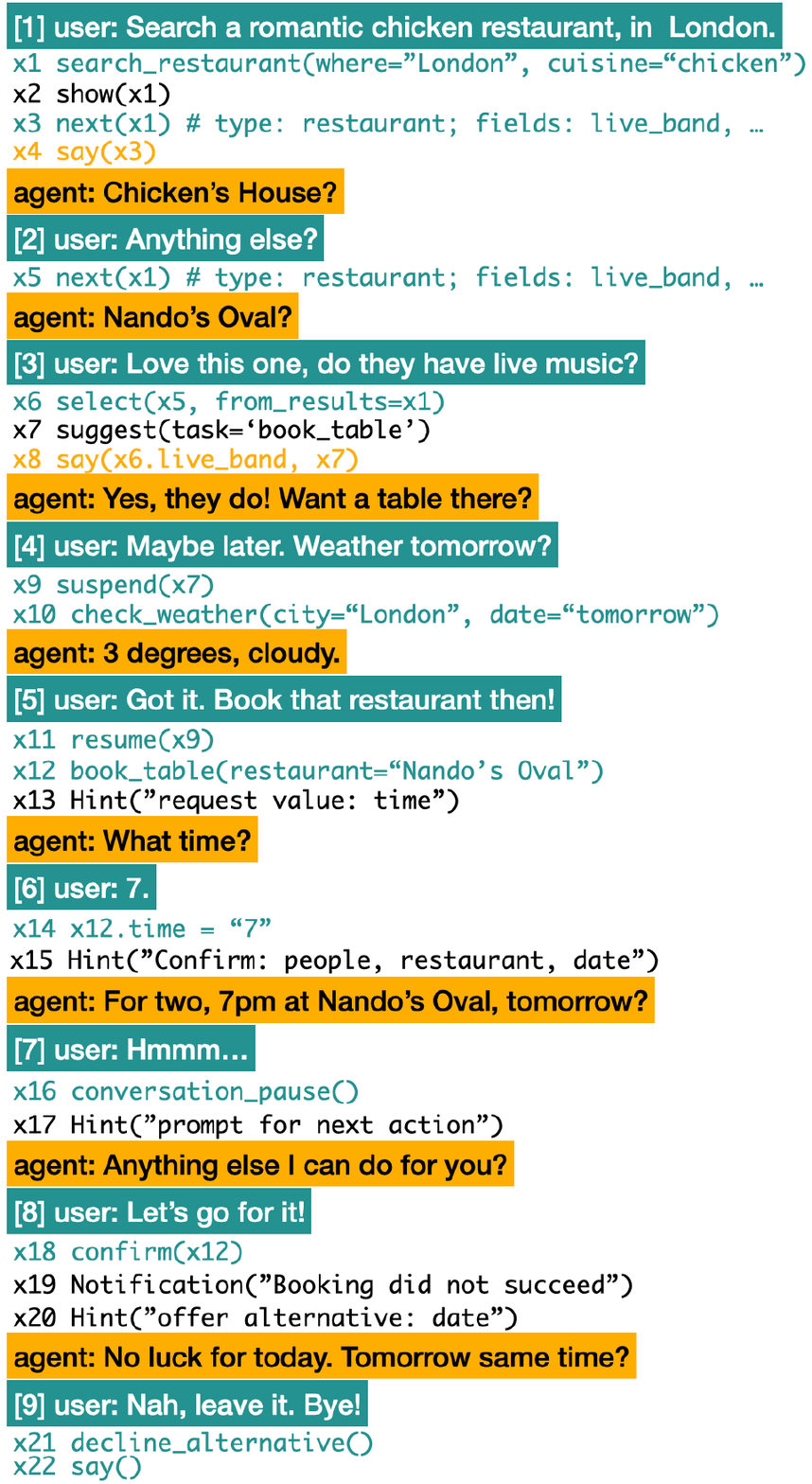}
    \centering
    \caption{Session transcript example. Natural Language Generation (NLG) calls shown only for turns 1 and 3.} 
    \label{fig:transcript}
\end{figure}

%% file: figures/user_actions.tex

\begin{table*}[t!]
\footnotesize
\centering
\setlength{\tabcolsep}{4pt}
\begin{tabular*}{\textwidth}{@{} p{1.5em} p{\dimexpr\textwidth-1.5em-8pt\relax} @{}}
\toprule

\rowcolor{gray!10}
\textbf{1} & \textbf{search API call} [e.g, \texttt{search\_restaurant}] \hfill \texttt{INFORM\_INTENT} \quad \textbf{[1]} \\
& \textbf{Generated when:} a user initiates a task which entails retrieving entities from a DB the agent can access (e.g., browsing). \\
& \textbf{Syntax:} keywords-only function call. Function name is a concatenation of service name and user intent. \\
& \textbf{Side effect:} updates API state with arguments provided by the user 
\\[1pt]

\rowcolor{gray!10}
\textbf{2} & \textbf{iteration} [\texttt{next}] \hfill \texttt{OFFER, REQ\_ALTS} \quad \textbf{[1], [2]} \\
& \textbf{Generated when:} search queries are fully specified or user browses through search results. \\
& \textbf{Syntax:} \textit{evar next(sapivar)} where \textit{sapivar} references a search API call. \\
& \textbf{Side effect:} fetches top search result from the database and stores it in \textit{evar}. \\[1pt]

\rowcolor{gray!10}
\textbf{3} & \textbf{selection} [\texttt{select}] \hfill \texttt{SELECT} \quad \textbf{[3]} \\
& \textbf{Generated when:} user selects an entity suggested by the agent (e.g., a restaurant). \\
& \textbf{Syntax:} \textit{select(evar, from=api)} where \textit{evar} is a search result and \textit{api} references a search API call. \\
& \textbf{Side effect:} API state updated with selected entity properties (e.g., \textit{restaurant="Nando's Oval"} at turn [3] in \Cref{fig:transcript}). \\[1pt]

\rowcolor{gray!10}
\textbf{4} & \textbf{suspend task} \texttt{[suspend]}\hfill \texttt{NEGATE\_INTENT} \quad \textbf{[3], [4]} \\
& \textbf{Generated when:} a user switches context, suspending a follow-up task suggested by the agent. \\
& \textbf{Syntax:} \textit{svar suspend(sugvar)} where \textit{sugvar} is a variable bound to an agent suggested task (e.g., x7 in \Cref{fig:transcript}). \\
& \textbf{Side effect:} \textit{svar} stores state relevant to the follow-up task (e.g., \textit{restaurant="Nando's Oval"} in \Cref{fig:transcript}). 
\\[1pt]

\rowcolor{gray!10}
\textbf{5} & \textbf{resume task} \texttt{[resume]} \hfill \texttt{INFORM\_INTENT} \quad \textbf{[5]} \\
& \textbf{Generated when:} a user continues a previously suspended task. \\
& \textbf{Syntax:} \textit{resume(svar)} where \textit{svar} is a variable referencing a suspended task. \\ 
& \textbf{Side effect:} Ensures resumed tasks (e.g., x12 in \Cref{fig:transcript}) are updated with state stored in \textit{svar}.\\[1pt]

\rowcolor{gray!10}
\textbf{6} & \textbf{transaction API call} [e.g., \texttt{book\_restaurant}] \hfill \texttt{INFORM\_INTENT} \quad \textbf{[5]} \\
& \textbf{Generated when:} a user initiates a task that changes their device state or that of an external service (e.g., bookings). \\
& \textbf{Syntax:} See \#1. \\
& \textbf{Side effect:} See \#1. \\[1pt]

\rowcolor{gray!10} 
\textbf{7} & \textbf{provide task constraints} \texttt{[--]}\hfill \texttt{INFORM} \quad \textbf{[6]} \\
& \textbf{Generated when:} a user provides task constraints or corrects slot values. \\
& \textbf{Syntax:} dot assignment: \textit{variable.field = value}, where \textit{variable} references a prior call and \textit{field} is an API slot.\\ 
& \textbf{Side effect:} updates referenced API state with slot values provided by the user.\\[1pt]

\rowcolor{gray!10}
\textbf{8} & \textbf{pause} \texttt{[conversation\_pause]} \hfill --- \quad \textbf{[7]} \\
& \textbf{Generated when:} detecting an out-of-scope utterance. \\
& \textbf{Syntax:} zero-arguments function call to \textit{conversation\_pause()}. \\ 
& \textbf{Side effect:} carries over the current API state to the next turn.\\[1pt]

\rowcolor{gray!10}
\textbf{9} & \textbf{confirm} \texttt{[confirm]} \hfill \texttt{AFFIRM} \quad \textbf{[8]} \\
& \textbf{Generated when:} a user confirms a transaction for execution. \\
& \textbf{Syntax:} \textit{confirm(apivar)} where \textit{apivar} references a transaction API call. \\ 
& \textbf{Side effect:} updates API state agent values (e.g., \textit{people=2, date="tomorrow", restaurant="Nando's Oval"} in \Cref{fig:transcript}). \\[1pt]

\rowcolor{gray!10}
\textbf{10} & \textbf{decline suggestions} \texttt{[decline\_alternative]} \hfill \texttt{NEGATE} \quad \textbf{[9]} \\
& \textbf{Generated when:} a user declines agent's suggestion to update task constraints following a transaction failure. \\
& \textbf{Syntax:} zero-arguments function call to \textit{decline\_alternative()}. \\
& \textbf{Side effect:} carries over the current API state to the next turn. \\
\bottomrule
\end{tabular*}
\caption{PyTOD user action sub-language summary. Shaded rows show equivalent DSTC8 dialogue acts \cite{DBLP:journals/corr/abs-2002-01359} and [digits] point to syntax examples and utterances grounded in these statements in \Cref{fig:transcript}.}
\label{tab:tod_syntax_final}

\end{table*}

%% file: figures/ss_prompts_v0.2.tex
\vspace{-0.5cm}
\begin{figure}[t]
    \centering
    \begin{subfigure}[b]{0.95\columnwidth}
        \centering
        \includegraphics[width=\columnwidth, trim={28mm 55mm 28mm 58mm}, clip]{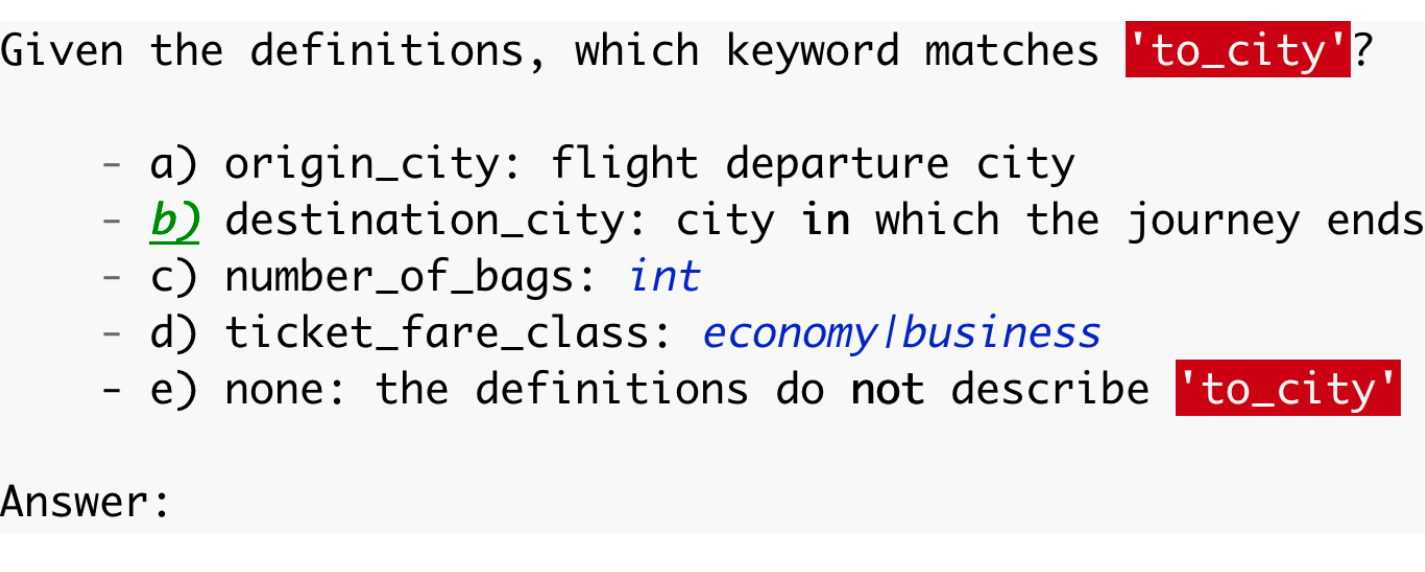}
        \vspace{-0.75cm}
        \caption{Unknown slot name. The \textit{\greenuline{SS output}} is converted to \textcolor{greenssprompt}{destination\_city}, replacing \textcolor{white}{\colorbox{customred}{to\_city}} in the AP output.}
        \label{fig:unk_slot_prompt}
    \end{subfigure}
    \vspace{0.25cm} 
    \begin{subfigure}[b]{0.95\columnwidth}
        \centering
        \includegraphics[width=\columnwidth, trim={11mm 50mm 10mm 45mm}, clip]{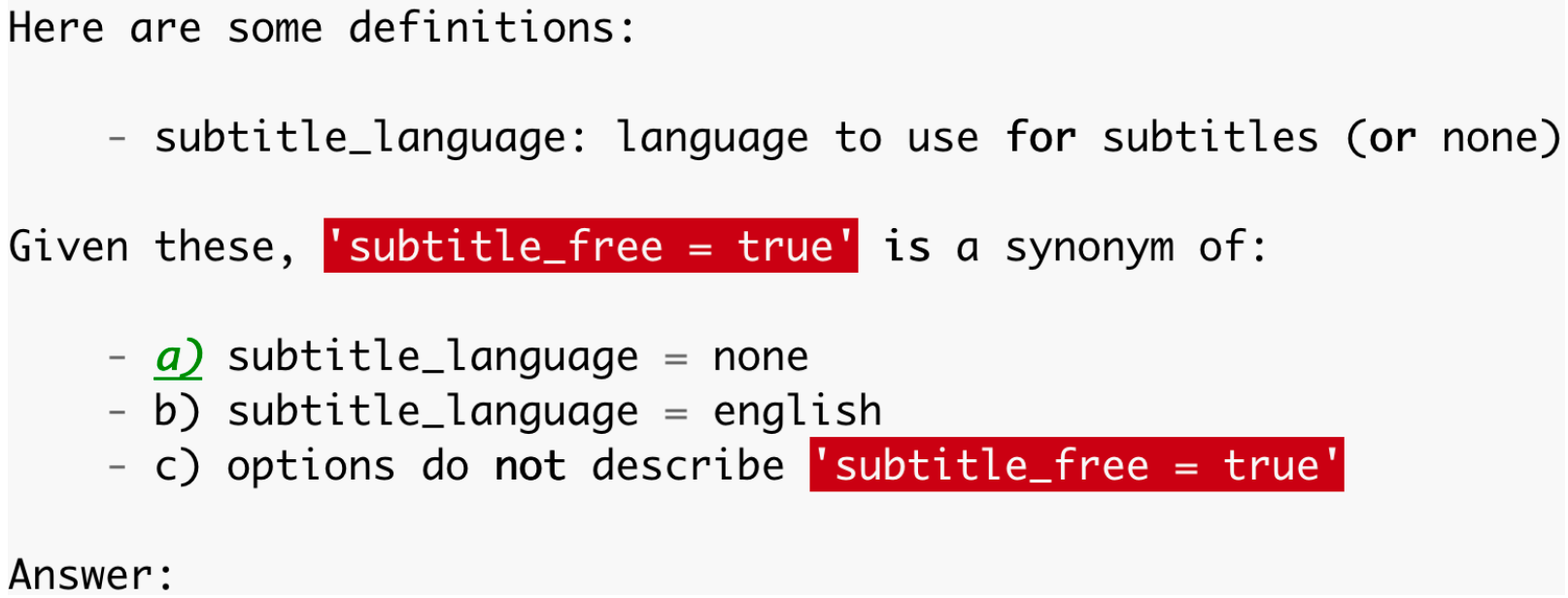}
        \vspace{-0.22cm}
        \caption{Unknown slot name (closed value). The \textit{\greenuline{SS output}} is converted to \textcolor{greenssprompt}{subtitle\_language = none}, replacing \textcolor{white}{\colorbox{customred}{subtitle\_free = true}} in the AP output.}
        \label{fig:unk_slot_cat_value_prompt}
    \end{subfigure}
    \begin{subfigure}[b]{0.95\columnwidth}
        \centering
        \includegraphics[width=\columnwidth, trim={25mm 71mm 25mm 72mm}, clip]{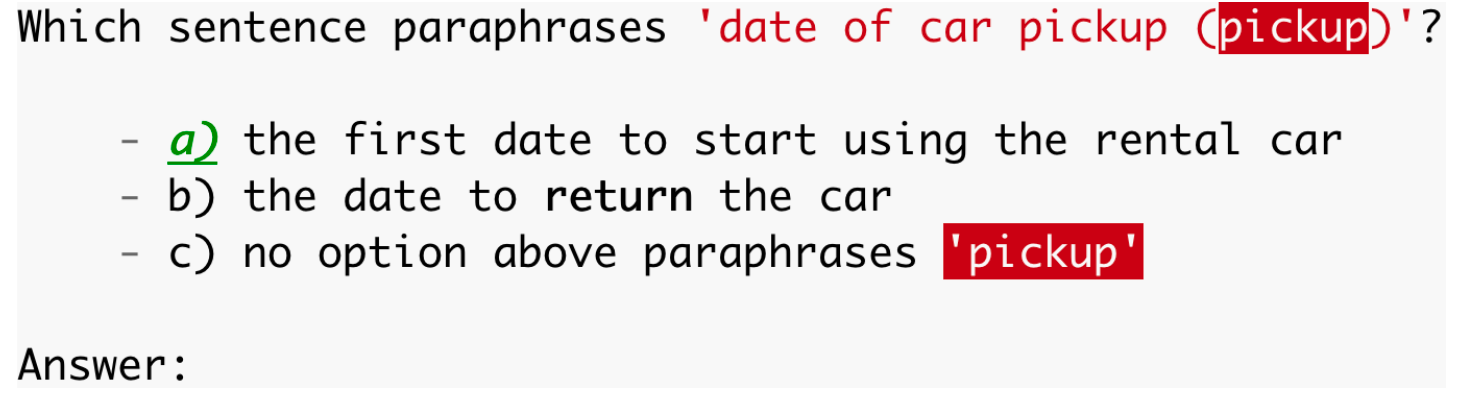}
        \vspace{-0.22cm}
        \caption{Memorised slot name. The \textit{\greenuline{SS output}} is converted to \textcolor{greenssprompt}{pickup\_date}, replacing \textcolor{white}{\colorbox{customred}{pickup}} in the AP output.
        \label{fig:memorisation_prompt}}
    \end{subfigure}
    \caption{Illustration of prompts generated by the SS to \textcolor{greenssprompt}{constrain} \textcolor{white}{\colorbox{customred}{AP generations}}  to the decoding schema.}
    \label{fig:ss_prompt}
\end{figure}

%% file: figures/ps_prompt.tex
\begin{figure}[h!]
\vspace{-0.2cm}
\includegraphics[width=0.95\linewidth, trim={23mm 40mm 23mm 40mm}, clip]{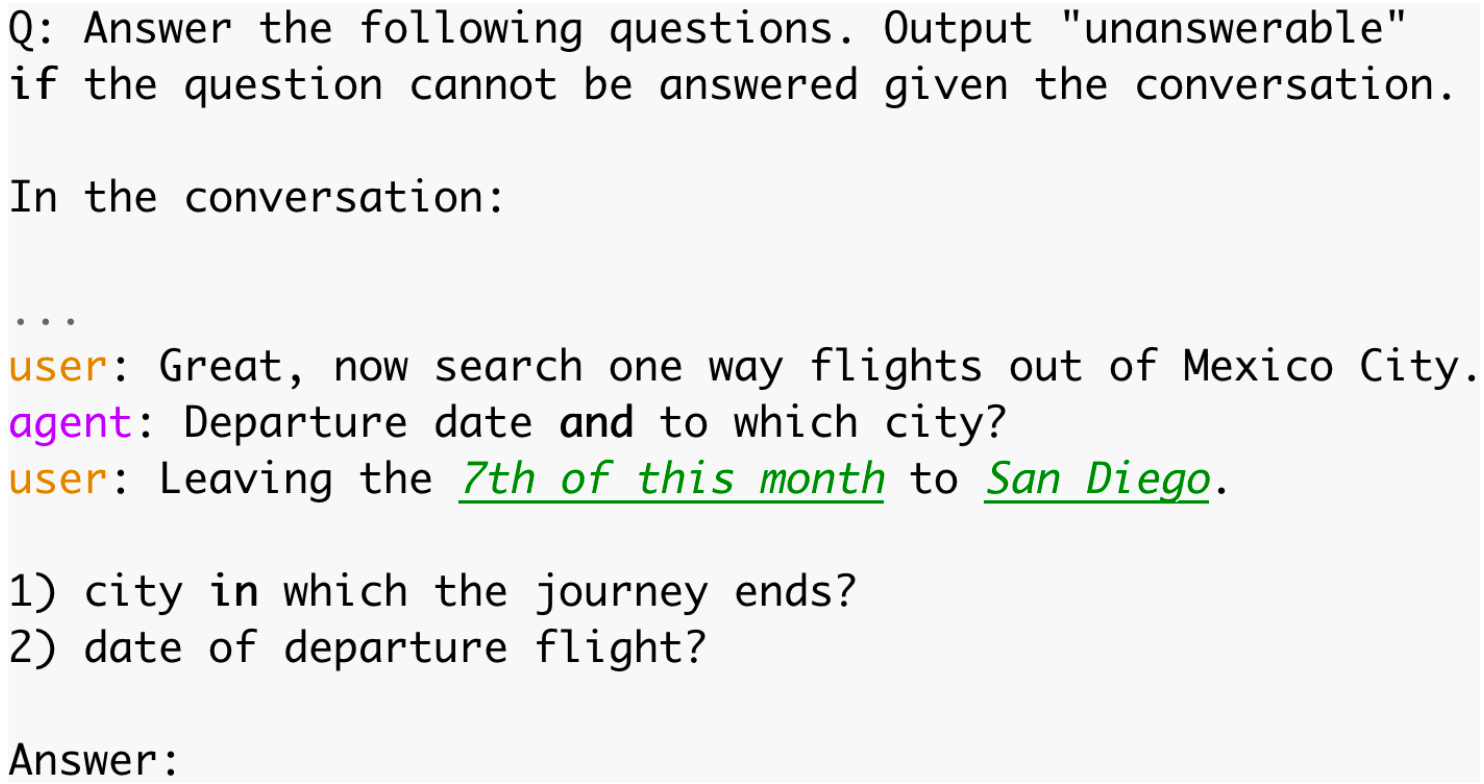}
    \centering
    \caption{Illustration of the prompt generated by the PS to handle AP omissions and semantic errors. The expected SS output is \textcolor{greenssprompt}{1) San Diego 2) 7th of this month}. The second answer is used to correct an omission by concatenating the constrained AP output in \Cref{fig:pytod} (x1.\textcolor{greenssprompt}{destination\_city} = "San Diego") with \textcolor{greenssprompt}{x1.departure\_date = "7th of this month"}.}
    \label{fig:ps_prompt}
    \vspace{-0.5cm}
\end{figure}

%% file: sections/experiments/v0.1.tex
\section{Experimental Setup}
\label{sec:experiments}
We evaluate PyTOD on schema-guided DST. While \citet{AnyTOD} also evaluate their system on next-action prediction (NAP), PyTOD is designed such that, given the correct dialogue state, it is guaranteed to take a correct action according to the dialogue policy. Hence, NAP is highly correlated with DST performance and does not provide additional insight into system behavior. Such turn-based evaluations often fail to reflect a system’s ability to satisfy complex user goals in real interactions \citep{takanobu,DBLP:journals/corr/abs-2412-01262}.
\subsection{Datasets and metrics}
\label{sec:datasets-metrics}
\textbf{Datasets} We experiment on SGD, which consists of $16,142$ training dialogues spanning $26$ \textit{service schemata}\footnote{A service exposes multiple APIs representing user intents.}. The test set comprises $4,201$ dialogues covering $21$ service schemata. SGD is designed to measure TOD generalization: among the $90$ distinct task sequences in the test set, $85.6\%$ involve a task grounded in a schema \textit{unseen} during training, which corresponds to $77\%$ of the dialogues.

\textbf{Metrics} We evaluate performance using \textit{joint goal accuracy (JGA)}, which measures the percentage of dialogue turns where all slot-value pairs are predicted correctly. To assess generalization, we report JGA separately for seen and unseen services. This distinction highlights each model’s ability to generalize to unseen slots and values and to correctly interpret API descriptions that were not encountered during training. We compute JGA using the official evaluator\footnote{Available at \href{https://bit.ly/3B7jD1c}{https://bit.ly/3B7jD1c}.} but extend the SGD state annotations with canonical slot values (\Cref{appendix:normalisation}). 

JGA evaluates context-to-state accuracy but if a slot correctly predicted in an earlier turn is later omitted when the state is re-estimated, the agent may prompt the user to needlessly repeat information, degrading interaction quality. We introduce C-JGA, a stricter metric that enforces state consistency: under C-JGA, a turn contributes to the JGA only if the state at all previous turns in the same task is jointly correct.
\subsection{Baselines}
\label{sec:baselines}
\uline{\textit{T5DST}} \cite{sgdx} jointly encodes the dialogue history and a slot description to generate the corresponding slot value. Each slot is processed independently, so one forward pass is required for each slot in the schema at every turn. \uline{\textit{D3ST}} \cite{d3st} predicts all active slot values in a single pass. \uline{\textit{SDT-Seq}} \cite{sdt} encode the dialogue history alongside a sample conversation and its state sequence to learn DST via in-context finetuning. \uline{PROMPTER} \cite{aksu-etal-2023-prompter} employ prefix-finetuning for zero-shot domain adaptation. \uline{\textit{LLaMA}} \cite{ldst} implements schema-guided DST with LLMs, and \uline{\textit{LDST}} \textit{(ibid.)} is a LLaMA finetuned with diverse, handcrafted prompts. \uline{\textit{AnyTOD}} is the SOTA E2E agent on SGD.
\subsection{State tracking with PyTOD}
\label{sec:pytod-dst}
\textbf{API retrieval} Consistent with all prior work on SGD, we assume service knowledge at each turn. Hence, PyTOD retrieves APIs from the AP header (\S \ref{sec:header}) and not the entire assistant schema. 

\textbf{Action parser} The AP is fine-tuned starting from FlanT5 \cite{flant5}, as described in \Cref{sec:pytod-impl}. Upon receiving a user utterance, the AP generates program statements which are validated by DM as described in \S \ref{sec:ap}. 

\textbf{Schema supervisor} The SS constrains the AP output based on the decoding schema (\Cref{fig:ss_prompt}). We use MQA prompts and PLM knowledge for \textit{zero-shot schema supervision} with FlanT5 (3B) \cite{flant5}, eliminating the need for slot paraphrase collection. This makes LLM-guided constrained decoding simple to implement. 

\textbf{Parser supervisor} The PS corrects slot omissions and semantic errors (\Cref{fig:ps_prompt}), with finetuning prompts constructed from the same dialogues as those used for AP finetuning. We perform \textit{multitask learning} for action parsing and parsing supervision. This allows PyTOD to function as a single model which performs corrections on DM request, with schema supervision handled by an off-the-shelf LLM. See fine-tuning details in \Cref{sec:pytod-impl}.

\textbf{Dialogue state prediction} The validated transcripts are executed line-by-line using \texttt{pytodlib}'s execution engine (\Cref{sec:exec-engine,appendix:execution_engine}), which changes the simulation state as described in \Cref{tab:tod_syntax_final}. At every user turn, the state of the dialogue is the state of the current API. The API state can be directly accessed from provided evaluation scripts and shares format with the SGD state annotations.

%% file: sections/results/v0.1.tex
\input{tables/main_results_v4}

\section{Main Result}
\label{sec:results}
TOD agents are deployed in resource-constrained settings, so we implement the AP and PS components using a small PLM and compare PyTOD against SOTA models of similar size and LLMs. \Cref{tab:jga_comparison} shows that PyTOD closely matches or outperforms all baselines, while also achieving higher consistency across turns. In particular, PyTOD (B) surpasses D3ST and T5DST by absolute margins of $5.6\%$ and $4.2\%$, respectively, with stronger performance on unseen services. PyTOD (L) achieves a similar improvement over D3ST (\#6\&11).

D3ST and T5DST re-estimate the state from scratch at each turn, meaning their JGA  increases if early errors are later corrected. In reality, however, these errors would alter the dialogue flow, so the JGA increases due to error recovery overestimate real-world performance. When adjusted for consistency, D3ST JGA drops sharply: $9\%$ (220M, \#3) and $8.8\%$ (780M, \#6). In contrast, PyTOD incrementally predicts user actions based on its \textit{predicted} past actions, resulting in smaller JGA drops of $4.1\%$ (B, \#10) and $3.8\%$ (L, \#11). These smaller decreases stem from two key factors. First, when users change goals mid-task, incorrect states may still be corrected. Second, parameters incorrectly copied from previous tasks at the start of a new task may be later overridden by their correct values.

PyTOD retrieves APIs from the prompt while the JGA of all other models is invariant to intent parsing errors, making the figures less sensitive to annotation errors (\S \ref{sec:err-analysis}) but less reflective of real-world TOD performance. PyTOD’s JGA therefore more accurately reflects practical deployment scenarios where intent errors impact interaction quality.

\textbf{Descriptions vs demonstrations} Beyond surpassing schema-guided baselines, PyTOD performs competitively with approaches requiring additional developer effort, such as SDT-Seq. The JGA gap between PyTOD and SDT-Seq is minimal: $0.7\%$ (B, \#5\&10) and $0.5\%$ (L, \#7\&11), with SDT-Seq performing better on seen domains. 

However, SDT-Seq requires developers to manually craft example demonstrations for each intent in addition to service schemas, increasing deployment overhead. Furthermore, SDT-Seq does not perform API retrieval, requiring an external intent detection model. Finally, PyTOD demonstrates substantially higher consistency, outperforming SDT-Seq by  $5.0\%$ (\#5\&10) and $4.2\%$ (\#7\&11) in C-JGA, reinforcing its stability in multi-turn interactions.

\textbf{LLM comparison} PyTOD compares favourably with LLMs. Compared to LLaMA, PyTOD (B) improves JGA by $1.5\%$ (\#8\&10) while being 32 times smaller. PyTOD (L) improves it by $6.9\%$ while being 10 times smaller (\#8\&11). These results show that automatic prompt generation given execution errors and policy feedback enables small language models to perform competitively with LLMs while operating at a fraction of a cost. LDST is trained with data generated with multiple prompt templates, outperforming our best model by $2.3\%$ (\#9\&11). We do not perform augmentation.

%% file: tables/main_results_v4.tex
\begin{table}[t]
\centering
\large
\resizebox{\columnwidth}{!}{%
\begin{tabular}{clccccc}
\toprule
\textbf{Size} & \textbf{Model} & \textbf{JGA } & \textbf{C-JGA}& \textbf{JGA (Seen)} & \textbf{JGA (Unseen)} & \textbf{\#} \\ 
\midrule
\multirow{5}{*}{220M} 
& PROMPTER  & -- & -- & -- & 49.4 & 1 \\
& T5DST & 72.6 & -- & 89.7 & 66.9 & 2 \\ 
& D3ST \textsuperscript{\textdagger} & 71.2 & 62.2 & 93.2 & 63.8 & 3 \\
& AnyTOD & -- & -- & 89.9 & 62.4 & 4 \\
& SDT-Seq \textsuperscript{\textdagger} & 77.5 & 68.7 & 93.5 & 72.2 & 5 \\
\midrule
\multirow{2}{*}{780M} 
& D3ST \textsuperscript{\textdagger} & 76.5 & 67.7 & 93.8 & 70.8 & 6 \\ 
& SDT-Seq \textsuperscript{\textdagger} & 82.7 & 74.2 & 94.1 & 78.9 & 7 \\ 
\midrule
\multirow{2}{*}{7B}
& LLaMA\textsuperscript{*}  & 75.3 & -- & -- & -- & 8 \\
& LDST\textsuperscript{*}  & 84.5 & -- & -- & -- & 9 \\
\midrule
220M & PyTOD (B) & 76.8 & 72.7 & 91.0 & 71.8 & 10 \\
780M & PyTOD (L) & 82.2 & 78.4 & 92.1 & 78.9 & 11 \\
\bottomrule
\end{tabular}%
}
\caption{PyTOD DST performance. Rows marked with \textdagger \space report the results of our replication study in \Cref{appendix:training-setup}. \textsuperscript{*} indicates training with prompt augmentation and evaluation with custom scripts as opposed to official evaluator.}
\label{tab:jga_comparison}
\end{table}

%% file: sections/analysis/v0.1.tex
\section{Analysis and Discussion}
\label{sec:analysis}
\subsection{Ablation study}
\label{sec:ablation}
\input{tables/ablation_study} 

\textbf{Parser supervisor} \Cref{tab:component-ablation} shows the PS improves PyTOD JGA by an average of $1.0\%$ (Base, \#1\&2) and $1.6\%$ (Large, \#5\&6). Analysing $210$ errors corrected by the PS in the best-performing PyTOD (Large) run ($82.6\%$ JGA) reveals that slot omissions occur more frequently than semantic errors ($61\%$ vs. $39\%$). The most common semantic error stems from confusion between similar slots (e.g., start date vs. end date). Sharing AP and PS parameters has a negligible effect on parsing accuracy (\#3\&4, \#7\&8), simplifying PyTOD deployment.

\input{tables/ss_size_analysis}
\input{tables/latency_analysis}
\textbf{Schema supervisor} While the AP generates slot names absent from the decoding schema for unseen services, the predicted slots often retain the correct semantics. For instance, the AP produces new slot names such as \textit{travel\_starts\_from} instead of \textit{journey\_starts\_from}. Other times, the AP outputs slots seen in a training schema implementing the same domain as an (unseen) decoding schema (e.g., outputs \textit{hotel\_name} instead of \textit{place\_name}), reflecting real-world challenges where TOD agents must support integration of services similar to the ones they have been trained on without further finetuning. The SS effectively mitigates both of these challenges, improving JGA by $11.4\%$ for PyTOD (Base, \Cref{tab:component-ablation}, \#2\&4) and $6.5\%$ for PyTOD (Large,  \#6\&8). Notably, the SS reduces system latency while improving performance: the 220M AP+SS system (\#2) runs $1.89$ times faster compared to the 780M AP (\#8) while being $1.7\%$ more accurate.

Beyond performance gains, LLM-constrained decoding simplifies deployment compared to grammar-based approaches. Unlike the latter, which require re-engineering to align with new backbone PLM tokenization schemes, LLM-based constrained decoding allows seamless AP updates.

\subsection{Does SS need to be an LLM?}
\label{sec:ss-size}

\Cref{tab:ss-size} shows that PLM-constrained decoding remains effective even as the SS size decreases: JGA drops by only $2.7\%$ (\#1\&5) for PyTOD (B) and $1.7\%$ (\#7\&11) for PyTOD (L) when reducing SS from 3B to 220M parameters, so LLM-guided constraint is feasible with small PLMs. The PS recovers some of the errors, improving JGA by an average of $3.55\%$ (PyTOD B) and $2.6\%$ (PyTOD L). Regardless of the AP, SS and PS configurations PyTOD consistently outperforms D3ST.

\textbf{Latency analysis} \Cref{tab:latency_ss_size} confirms that SS size has a minimal impact on PyTOD latency. Expanding SS from 220M to 3B increases latency by only $11\%$ for PyTOD (Base) and just $2\%$ for PyTOD (Large). This is expected since the SS prompts are short and the MQA formatting enables  SS to constrain decoding with a single token. Most of the latency increase stems from on-demand SS model loading, which can be optimized by keeping SS in memory at the cost of a higher memory footprint.

\subsection{Error analysis}
\label{sec:err-analysis} 
\textbf{Seen services} An analysis of $20$ dialogues from each of the $3$ services with JGA below SDT-Seq average, reveals that \textit{RideSharing\_2 (RS\_2)} and \textit{Movies\_1 (MOV\_1)} contribute most to the discrepancy. For \textit{RS\_2}, $70\%$ of the  errors involve incorrect slot values: the system consistently misinterprets requests like \textit{cheapest ride} as \textit{regular} instead of \textit{pool}, suggesting that PyTOD could benefit by implementing the AP and PS with a PLM that has stronger world knowledge. For \textit{MOV\_1}, value errors are primarily \textit{span errors}, where the model fails to capture the full movie title or crosses span boundaries. Notably, annotation errors where the annotated slot values are absent from the user utterance predominate in \textit{Travel\_1} samples.
\input{tables/error_analysis_v0.8}

\textbf{Unseen services} Annotation errors significantly impact unseen service performance, particularly in \textit{Trains\_1}, where intent paraphrase errors (\Cref{appendix:annotation}) prevent PyTOD from tracking state due to its reliance on retrieved train schedules. When search tasks succeed, \textit{copy errors} - where PyTOD fails to propagate query parameters (e.g., \textit{album} from \textit{LookupMusic} to \textit{PlayMedia}) reduce JGA in \textit{Movie\_3} and \textit{RentalCars\_3}. For example, while \textit{GetCarsAvailable} succeeds $90\%$ of the time, \textit{ReserveCar} succeeds in just $26.23\%$ of cases. This highlights systematic failures which could be addressed through targeted improvements.

In \textit{Messaging\_1}, copy errors stem from incorrect co-reference resolution: the \textit{location} slot is resolved to cities instead of addresses due to training schema bias: \textit{location} always describes cities in training. TOD agents that support multiple services are prone to similar errors when concept names differ across training schemas, and fine-tuning alone can lead to overfitting to schema-specific naming conventions. Future work will explore enhanced supervision mechanisms to mitigate these biases.

%% file: tables/ablation_study.tex
\begin{table}[t]
\centering
\resizebox{\columnwidth}{!}{%
\begin{tabular}{clcccccc}
\toprule
\textbf{Size} & \textbf{PS} & \textbf{SS} & \textbf{Multitask} & \textbf{JGA} & \textbf{JGA Seen} & \textbf{JGA Unseen} & \textbf{\#} \\ 
\midrule
\multirow{4}{*}{220M} 
& \textcolor{green}{\ding{51}} & \textcolor{green}{\ding{51}} & \textcolor{green}{\ding{51}} & 76.8 & 91.0 & 72.1 & 1 \\
& \textcolor{red}{\ding{55}} & \textcolor{green}{\ding{51}} & \textcolor{red}{\ding{55}} & 75.8 & 90.1 & 71.0 & 2 \\
& \textcolor{red}{\ding{55}} & \textcolor{red}{\ding{55}} & \textcolor{green}{\ding{51}} & 64.2 & 88.2 & 56.2 & 3 \\
& \textcolor{red}{\ding{55}} & \textcolor{red}{\ding{55}} & \textcolor{red}{\ding{55}} & 64.4 & 88.8 & 56.3 & 4 \\
\midrule
\multirow{4}{*}{780M} 
& \textcolor{green}{\ding{51}} & \textcolor{green}{\ding{51}} & \textcolor{green}{\ding{51}} & 82.2 & 92.1 & 78.9 & 5 \\
& \textcolor{red}{\ding{55}} & \textcolor{green}{\ding{51}} & \textcolor{red}{\ding{55}} & 80.6 & 91.5 & 77.0 & 6 \\
& \textcolor{red}{\ding{55}} & \textcolor{red}{\ding{55}} & \textcolor{green}{\ding{51}} & 74.6 & 90.8 & 69.2 & 7 \\
& \textcolor{red}{\ding{55}} & \textcolor{red}{\ding{55}} & \textcolor{red}{\ding{55}} & 74.1 & 90.4 & 68.6 & 8 \\
\bottomrule
\end{tabular}%
}
\caption{Contribution of parser supervisor (PS) and schema supervisor (SS) to PyTOD performance. \textbf{Multitask} indicates joint training of the action parsing and parsing supervision tasks. Rows 10\&11 from \Cref{tab:jga_comparison} are repeated in rows 1\&5 to facilitate comparisons.}
\label{tab:component-ablation}
\vspace{-0.4cm}
\end{table}

%% file: tables/ss_size_analysis.tex
\begin{table}[t]
\centering
\resizebox{0.9\columnwidth}{!}{%
\begin{tabular}{clccccc}
\toprule
\textbf{AP Size} & \textbf{SS Size} & \textbf{PS} & \textbf{JGA} & \textbf{JGA Seen} & \textbf{JGA Unseen} & \textbf{\#} \\ 
\midrule
\multirow{6}{*}{220M} 
& \multirow{2}{*}{3B} 
& \textcolor{green}{\ding{51}} & 76.8 \deemph{(+5.6\%)} & 91.0 & 72.1 & 1 \\ 
& & \textcolor{red}{\ding{55}} & 75.8 \deemph{(+4.6\%)} & 90.1 & 71.0 & 2 \\ 
\cmidrule(lr){2-7}
& \multirow{2}{*}{780M} 
& \textcolor{green}{\ding{51}} & 75.1 \deemph{(+3.9\%)} & 89.9 & 70.2 & 3 \\ 
& & \textcolor{red}{\ding{55}} & 71.3 \deemph{(+0.1\%)} & 88.8 & 65.4 & 4 \\ 
\cmidrule(lr){2-7}
& \multirow{2}{*}{220M} 
& \textcolor{green}{\ding{51}} & 74.1 \deemph{(+2.9\%)} & 89.7 & 68.8 & 5 \\ 
& & \textcolor{red}{\ding{55}} & 70.8 \deemph{(-0.4\%)} & 89.0 & 64.7 & 6 \\ 
\midrule
\multirow{6}{*}{780M} 
& \multirow{2}{*}{3B} 
& \textcolor{green}{\ding{51}} & 82.2 \deemph{(+5.7\%)}& 92.1 & 78.9 & 7 \\ 
& & \textcolor{red}{\ding{55}} & 80.6 \deemph{(+4.1\%)} & 91.5 & 77.0 & 8 \\ 
\cmidrule(lr){2-7}
& \multirow{2}{*}{780M} 
& \textcolor{green}{\ding{51}} & 81.2 \deemph{(+4.7\%)} & 91.2 & 77.9 & 9 \\ 
& & \textcolor{red}{\ding{55}} & 78.6 \deemph{(+2.1\%)} & 90.3 & 74.7 & 10 \\ 
\cmidrule(lr){2-7}
& \multirow{2}{*}{220M} 
& \textcolor{green}{\ding{51}} & 80.5  \deemph{(+4.0\%)}& 91.2 & 76.9 & 11 \\ 
& & \textcolor{red}{\ding{55}} & 77.9 \deemph{(+1.4\%)} & 90.3 & 73.7 & 12 \\ 
\bottomrule
\end{tabular}%
}
\caption{PyTOD performance as function of SS size. Checkmarks (\textcolor{green}{\ding{51}}) and  crosses (\textcolor{red}{\ding{55}}) indicate PS presence or absence, respectively. Rows 10 and 11 from \Cref{tab:jga_comparison} are repeated in rows 1 and 7, for easy comparison. Bracketed numbers represent absolute improvements with respect to our implementation of D3ST (\Cref{tab:jga_comparison}, \#3\&6)}
\label{tab:ss-size}
\vspace{-0.6cm}
\end{table}

%% file: tables/latency_analysis.tex
\begin{table}[t]
\centering
\resizebox{0.9\columnwidth}{!}{%
\begin{tabular}{cccccccc}
\toprule
\textbf{AP Size} & \textbf{SS Size} & \textbf{Samples per sec}  & \textbf{Runtime (sec)} & \textbf{Relative Latency} & \textbf{\#} \\ 
\midrule
\multirow{4}{*}{220M} 
& - & 9.110 & 5875.12 & 1.00 & 1 \\ 
& 220M & 6.816 & 7919.79 & 1.35 & 2 \\ 
& 780M & 6.766 & 7946.26 & 1.35 & 3 \\ 
& 3B & 6.323 & 8588.64 & 1.46 & 4 \\ 
\midrule
\multirow{4}{*}{780M} 
& - & 3.331 & 16269.70 & 1.00 & 5 \\ 
& 220M & 2.971 & 18300.93 & 1.12 & 6 \\ 
& 780M & 2.964 & 18341.15 & 1.13 & 7 \\ 
& 3B & 2.903 & 18771.32 & 1.15 & 8 \\ 
\bottomrule
\end{tabular}%
}
\caption{Latency analysis. Numbers reported are for models without PS (marked with \textcolor{red}{\ding{55}} in \Cref{tab:ss-size}). Each figure is an average of three runs of making predictions for  $\approx53\mathrm{k}$ test set turns on an NVIDIA A100 GPU.}
\label{tab:latency_ss_size}
\vspace{-0.6cm}
\end{table}

%% file: tables/error_analysis_v0.8.tex
\newcommand{\errval}[2]{#1\deemph{(#2)}}  

\begin{table}[t]
\centering

\setlength{\tabcolsep}{3pt}

\resizebox{\columnwidth}{!}{%
\begin{tabular}{@{}l@{\hspace{2pt}}  
                c@{\hspace{4pt}} c@{\hspace{4pt}} c@{\hspace{4pt}}%
                c@{\hspace{4pt}} c@{\hspace{4pt}} c@{\hspace{4pt}}%
                c@{\hspace{4pt}} c@{}}
\toprule

\multirow{3}{*}{%
  \parbox[c]{0.9cm}{\centering \textbf{Error}\\\textbf{Type}}%
} 
 & \textbf{RideSharing\_2} & \textbf{Movies\_1} & \textbf{Travel\_1}
 & \textbf{RentalCars\_3}  & \textbf{Trains\_1}
 & \textbf{Music\_3}       & \textbf{Messaging\_1}
 & \textbf{Total} \\

\cmidrule(lr){2-8}
  & \parbox[c]{1.4cm}{\centering 78.2\\\deemph{(-13.9\%)}} 
  & \parbox[c]{1.4cm}{\centering 80.7\\\deemph{(-11.4\%)}} 
  & \parbox[c]{1.4cm}{\centering 92.5\\\deemph{(+0.4\%)}} 
  & \parbox[c]{1.4cm}{\centering 55.3\\\deemph{(-23.6\%)}} 
  & \parbox[c]{1.4cm}{\centering 60.2\\\deemph{(-18.7\%)}} 
  & \parbox[c]{1.4cm}{\centering 64.0\\\deemph{(-14.9\%)}} 
  & \parbox[c]{1.4cm}{\centering 65.7\\\deemph{(-13.2\%)}} 
  & -- \\

  & \textcolor{green}{\ding{51}} & \textcolor{green}{\ding{51}} & \textcolor{green}{\ding{51}}
  & \textcolor{red}{\ding{55}}   & \textcolor{red}{\ding{55}}
  & \textcolor{red}{\ding{55}}   & \textcolor{red}{\ding{55}}
  & -- \\
\midrule

\textbf{Missed} 
 & \errval{15.0}{3} & \errval{5.9}{2} & \errval{9.5}{2} 
 & \errval{3.7}{1}  & \errval{9.4}{3} & \errval{20.0}{5} & -- 
 & 16\deemph{(9.2)} \\

\textbf{Value}
 & \errval{70.0}{14} & \errval{32.4}{11} & --
 & \errval{22.2}{6}  & \errval{9.4}{3} & \errval{36.0}{9} & -- 
 & 43\deemph{(24.9)} \\

\textbf{Copy}             
 & \errval{15.0}{3} & \errval{29.4}{10} & \errval{4.8}{1}
 & \errval{48.1}{13} & \errval{21.9}{7} & \errval{28.0}{7} & \errval{78.6}{11} 
 & 52\deemph{(30.1)} \\

\textbf{Annot.}       
 & -- & \errval{17.6}{6}  & \errval{71.4}{15}
 & \errval{14.8}{4} & \errval{40.6}{13} & \errval{16.0}{4} & \errval{21.4}{3} 
 & 45\deemph{(26.0)} \\

\textbf{Intent}           
 & -- & --   & --
 & -- & \errval{9.4}{3}   & --  & -- 
 & 3\deemph{(1.7)} \\

\textbf{Other}            
 & -- & \errval{14.7}{5}  & \errval{14.3}{3}
 & \errval{11.1}{3} & \errval{9.4}{3}   & -- & --  
 & 14\deemph{(8.1)} \\

\midrule
\textbf{Total}
 & 20 & 34 & 21
 & 27 & 32 & 25 & 14
 & 173\deemph{(100)} \\
\bottomrule
\end{tabular}
} 

\caption{Errors of the best PyTOD (L) run, for seen (\textcolor{green}{\ding{51}}) and unseen (\textcolor{red}{\ding{55}}) services. Second row shows service JGA and its absolute deviation with respect to the average seen/unseen JGA in \Cref{tab:jga_comparison} (row 11). Percentages indicate the fraction of total service errors and (raw counts) are shown. \textbf{Missed}=omitted slot, \textbf{Value}=incorrect value, \textbf{Copy}=copied incorrect value/did not copy value, \textbf{Annot.}=annotation err., \textbf{Intent}=incorrect intent.}
\label{tab:error_analysis}
\vspace{-0.65cm}
\end{table}

%% file: sections/related_work/v0.1.tex
\section{Related Work}
\label{sec:related_work}
\textbf{Few- and zero-shot DST} Our approach and baselines (\S \ref{sec:baselines}) extend prior transfer learning work on cross-domain DST generalization via schema descriptions \cite{descriptions-cross-domain-lin} and QA tasks \cite{qa-cross-domain-lin}. Recently, large-scale proprietary LLMs (e.g., ChatGPT \cite{ChatGPT}) have shown strong DST performance with no \cite{milica-chatgpt} or few \cite[\textit{inter alia}]{function-calling-dst, icdst, ldst, json-and-rules-near-scoop} training dialogues. Like PyTOD, these approaches predict state updates, represented in JSON format \cite{json-and-rules-near-scoop}, as SQL queries \cite{icdst}, function calls \cite{other-function-calling, function-calling-dst} or code \cite{python-syntax-dst-llm}. Unlike PyTOD, they are not policy-guided nor do they operate in a simulated environment; instead they track slots from user and agent utterances - a limitation that degrades performance \cite{json-and-rules-near-scoop}.  

These methods focus on data-efficient DST, relying on LLMs at inference. Recent works \cite{andybro, d0t} use LLMs for data generation, improving generalization without costly per-turn inference and addressing concerns over cost, resource availability, and privacy \cite{milica-chatgpt, ldst}. Distillation \cite{seanie, capid} and data augmentation \cite{ldst, function-calling-dst} provide alternatives, though they still depend on billion-scale models. To improve efficiency, subsequent works \cite{correction-lm, orchestrallm} employ self-correction \cite{correctable-dst, amendable-dst} to maintain accuracy. PyTOD targets stricter resource constraints, achieving accurate and consistent DST with models an order of magnitude smaller. It eschews corrector fine-tuning or feedback generation, using readily available execution feedback and simple QA prompts for zero-shot correction with one token. Its policy- and environment-driven corrections avoid per-turn supervision, improving efficiency.

\textbf{TOD agents} Transfer learning \cite{AnyTOD} and prompting \cite[\textit{inter alia}]{dialog2API, SGPDST, directional-stimulus, icl-hints} also apply to TOD agent development. \texttt{pytodlib} provides challenges beyond MultiWOZ through policy and ontology complexity \cite{hudecek-dusek-2023-large, correction-lm, selfexplanation}, providing a testbed for advancing TOD agents while addressing the dearth of conversational tool-use corpora \cite{toolsandbox}.

%% file: appendices/simulation.tex
\section{\texttt{pytodlib}}
\label{appendix:pytod-library-overview}
We describe key \texttt{pytodlib} features, encouraging interested readers to consult our software release.
\subsection{Service APIs}
\label{appendix:apis-sim}
\Cref{fig:sgd_api} shows a sample implementation of the \texttt{Buses\_3} service from SGD. It provides two APIs: (1) \texttt{FindBus}, a \textit{search} or \textit{query} API (\Cref{fig:sgd_api_search}) that enables users to query a bus schedule database using natural language, and (2) \texttt{BuyBusTicket}, a \textit{transactional} API (\Cref{fig:sgd_api_transaction}) that allows users to purchase a ticket for an itinerary proposed by the TOD agent based on  \texttt{FindBus} search results or by specifying  ticket details directly.
\input{figures/sgd_api_v1}

The APIs define slots as class variables implementing the \textit{descriptor protocol}\footnote{See \href{https://bit.ly/4hLB4aU}{https://bit.ly/4hLB4aU}.}. This enables advanced functionality such as \textit{execution error feedback}: setting an undefined attribute on an API could return a string describing the error instead of raising an exception. Another example is \textit{type coercion}: descriptors cast slot values to the data types specified in the schema, and may be configured to provide natural language feedback if the conversion fails, providing yet another opportunity for generating error-correction prompts. We leave these experiments to future work. 

\subsection{Policy simulation}
\label{appendix:pytod-dialogue-policy}
The SGD conversations are generated by sampling from a \textit{policy graph} \cite{SGD, stardataset}, which outlines the intended flow of a dialogue. Both search and transactional APIs require zero or more specific slots to function. To provide them, the system processes the initial user turn and takes actions to elicit missing slot values. Once all required slots are filled, search APIs can construct a valid database query, while transactional APIs execute  an external service call (e.g., to a ticket booking service).
\input{figures/execution_engine}

Slot-filling is abstracted in the \texttt{Command} interface\footnote{See \texttt{src/simulation/command.py}}, and the \texttt{SearchCommand}\footnote{See \texttt{src/simulation/search\_command.py}} interface (implementing search use cases) and the \texttt{ConfirmedCommand}\footnote{\texttt{src/simulation/confirmed\_command.py}} interface (implementing transaction-specific policies) inherit. All concrete service implementations (e.g., \texttt{Buses\_3}) extend the search and/or transaction interfaces. Upon execution, the interfaces return system actions, such as \texttt{show}, \texttt{perform}, \texttt{Hint} and \texttt{Notification}. For example, if a user says, \textit{I need a bus from London to Manchester.}, executing \texttt{FindBus} returns a system action \texttt{Hint(request value: departure\_date)}\footnote{The other arguments in \Cref{fig:sgd_api} are optional, so are not requested by the system.}, which an agent can verbalise to ask for the missing constraints. We encourage the interested reader to consult the detailed examples in \Cref{tab:dialogue} and revisit  \Cref{fig:pytod} in \S \ref{sec:intro}, \Cref{fig:transcript} in \S \ref{sec:transcript}  and \Cref{fig:transcript_details} in \Cref{appendix:ap-prompt} to deepen their understanding of the agent-environment interaction protocol \texttt{pytodlib} implements.

Unlike a majority of examples in the STARv2 \cite{AnyTOD} dataset, where users provide constraints only when prompted and the system follows a purely deterministic behaviour, SGD conversations follow a complex policy graph, that frequently includes user-initiated constraint specification and extends beyond the standard slot-filling approach in MultiWOZ. We encourage readers to explore the documentation of the \texttt{simulation} package in our software supplementary.

\subsection{Execution engine} 
\label{appendix:execution_engine}
The execution engine converts the DM’s output into a {Python} object. \Cref{fig:execution-engine} illustrates how a program statement is interpreted as an API object implemented by a service (e.g., \texttt{FindBus} in \Cref{fig:sgd_api_search}). The execution process begins by parsing the statement into an AST (line 24). If the tree matches a function call signature (line 26), a conversion function first extracts the function name and arguments (lines 6–10). If the function corresponds to an API implemented by the schema (line 14), the engine retrieves the command from a registry, instantiates the appropriate object (e.g., \texttt{FindBus}), and assigns slot values (lines 15–18). For user or system actions, the function name is returned (lines 20–22), and object instantiation is handled downstream. Interested readers should consult the documentation of the \texttt{execution} package in our software release for further details. 

%% file: figures/sgd_api_v1.tex
\begin{figure}[htpb]
    \centering
    \begin{subfigure}[t]{\linewidth}
        \centering
        \includegraphics[width=0.9\linewidth, trim={20mm 60mm 20mm 60mm}, clip]{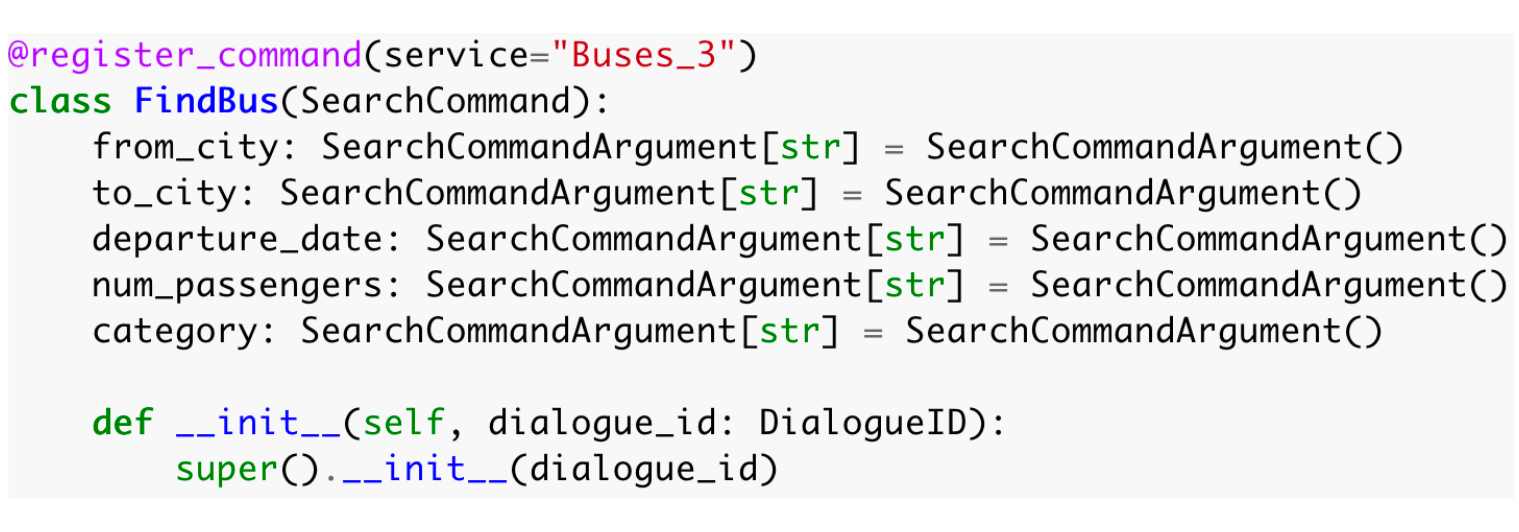}
        \caption{Search API.}
        \label{fig:sgd_api_search}
    \end{subfigure}
    \begin{subfigure}[t]{\linewidth}
        \centering
        \includegraphics[width=0.9\linewidth, trim={0mm 60mm 0mm 60mm}, clip]{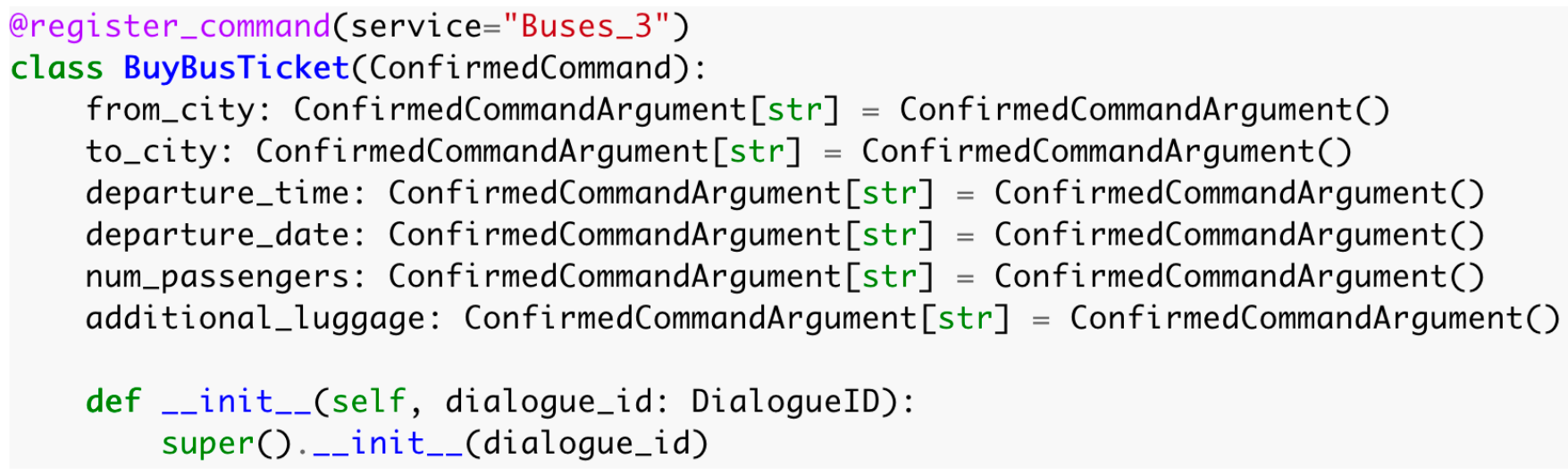}
        \caption{Transactional API.}
        \label{fig:sgd_api_transaction}
    \end{subfigure}
    \caption{Implementation of the SGD {Buses\_3} service}
    \label{fig:sgd_api}
\end{figure}

%% file: figures/execution_engine.tex
\begin{figure}[t]
\includegraphics[width=\linewidth, trim={25mm 0mm 25mm 0mm}, clip]{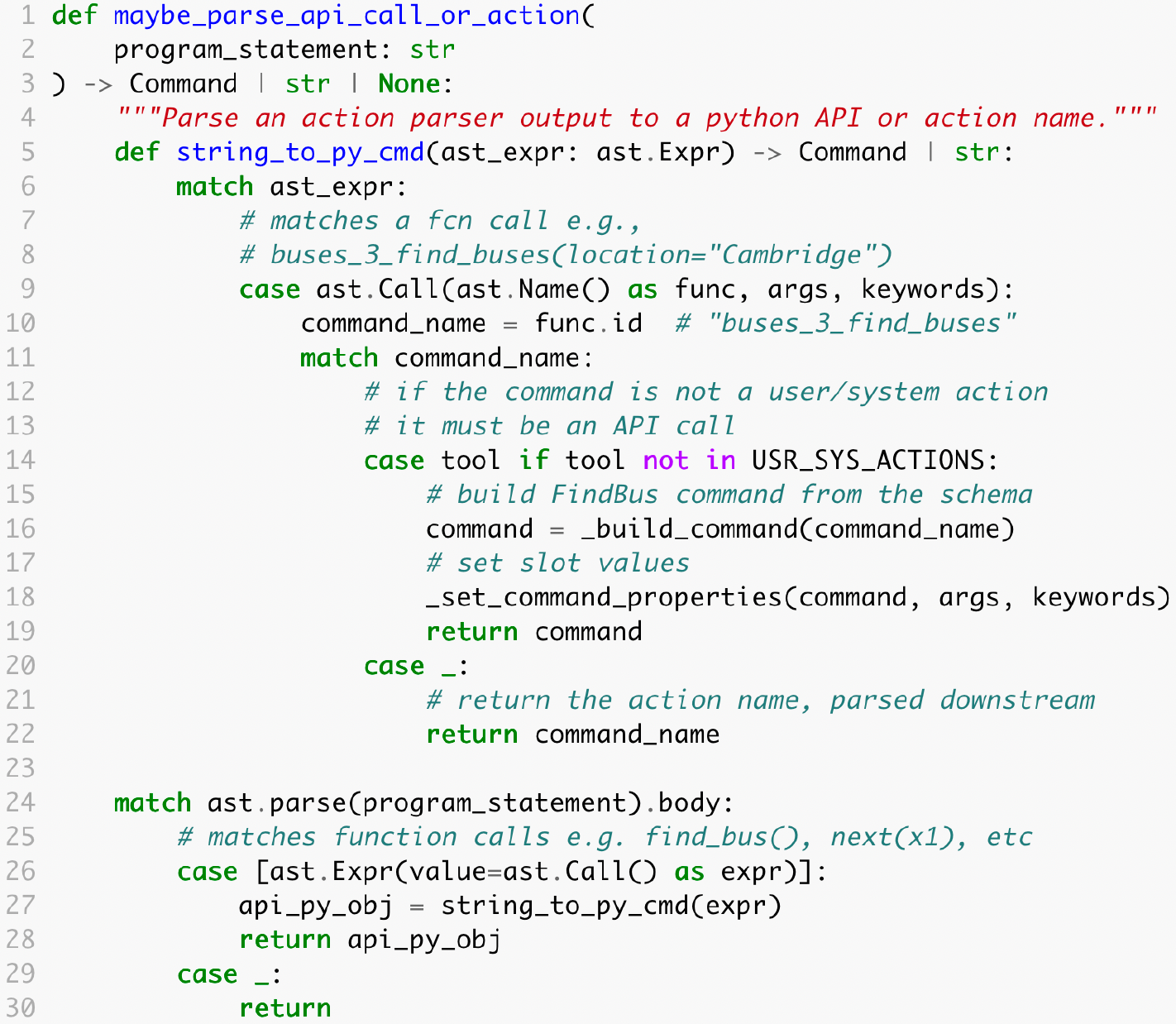}
    \centering
    \caption{Sample execution engine code, showing how a program statement is interpreted as a \texttt{python} object.}
    \label{fig:execution-engine}
\end{figure}

%% file: appendices/prompt_details_v0.3.tex
\input{figures/header_prompt}

\section{Prompts}
\label{appendix:prompting}
\subsection{Action parser}
\label{appendix:ap-prompt}
\subsubsection{Header} 
\label{appendix:header_prompt}

\Cref{fig:action_parser_prompt} presents a sample AP prompt header. If no task has been completed, the prompt begins with task instructions (\Cref{fig:task_instruction}). Once a task is completed, a summary of the task and the returned entity is prepended before the task instructions (\Cref{fig:task_stack}). The entity definitions include a docstring instructing the model to copy relevant argument values to the parameters of subsequent API calls.
\subsubsection{Session transcript} 
\label{appendix:session_transcript_prompt}
This section provides additional insight regarding the generality of PyTOD programs and system actions.
\input{figures/context_dependent_instructions} 

\textbf{Handling compositional utterances} \Cref{fig:transcript_details} demonstrates how PyTOD can handle compositional user utterances requiring nested function calls. At turn 9, the user declines alternative restaurant booking arrangements proposed by the agent, requesting directions to their next meeting. To resolve the meeting location, the agent must retrieve user's schedule from their calendar, sort it and then read the location of the next event (x29 - x31). 

Such utterances cannot represented by AnyTOD or other state-of-the-art TOD agents and DST models, which are limited to parsing slot-value pairs from user and agent utterances.
\input{figures/transcript_details}

\textbf{System actions} \S \ref{sec:transcript} introduced the \texttt{perform} statement as a marker of successful API execution. An example of its syntax and use is shown in \Cref{fig:transcript_details} at turn 9. By passing the reference of the variable bound to this statement to the subsequent \texttt{say} call, the NLG module can generate an utterance which informs the user their task executed successfully. This is equivalent to a \texttt{NOTIFY\_SUCCESS} in the DSTC8 dialogue act tagset.

See \Cref{tab:dialogue} for a summary of all PyTOD system and user actions along with further examples.

\subsubsection{Context-dependent instructions}
\label{appendix:context_dependent_instructions_prompt}
Context dependent instructions are formatted as \textcolor{blue}{{developer}:} turns, appear in the prompt after an iteration (i.e., \texttt{next)} or confirmation instructions. In the former case (\Cref{fig:cd_instr_qa}), they provide an itemized list of entity properties that the user may request along with their natural language descriptions. A brief instruction precedes this list, prompting the language model to invoke the \texttt{say} routine to communicate the requested information to the user. In the latter case (\Cref{fig:cd_instr_policy}), additional system policy instructions relevant to state tracking may be included. For instance, line 6-9 in \Cref{fig:cd_instr_policy} illustrate how PyTOD can be guided to correctly parse API parameters following an API calling error.

\subsection{Dialogue manager}
\label{appendix:dm-prompt}
\subsubsection{Schema supervisor}
\label{appendix:ss-templates}
\Cref{fig:ss_templates} shows the SS prompt generation templates. A common input to these is \texttt{slot\_schemas}, a list of dictionaries containing the names, descriptions, data type and possible values (for categorical slots) for the active service. 
\input{figures/ss_templates}
\input{figures/ps_templates}
These are formatted by developer-defined filters (e.g., \texttt{slot\_definition\_formatter}). The template for constraining the value of a categorical slot to one of the values listed in the schema (\Cref{fig:unk_cat_value}) is a special case of the template for constraining unknown slot names with categorical values (\Cref{fig:unk_arg_closed_val}): only one slot definition corresponding to the one predicted by the AP is displayed and the \texttt{none} option is removed to ensure the output is one of the values enumerated in the schema.

\subsection{Parser supervisor}
\label{appendix:ps-template}
\Cref{fig:ps_template} shows the PS prompt generation template. The DM filters the transcript to extract \textcolor{customorange}{user} and \textcolor{custompurple}{agent} turns relevant to the current task, ignoring previous tasks which are not relevant for predicting the current slot value. These turns are processed by the \texttt{conversation\_formatter} filter, which preprends the conversation role to the utterance. The schemata of the slots requested at the previous turned are passed to the template as \texttt{slot\_list}. The \texttt{question\_formatter} filter the formats the slot descriptions by lowercasing them and appending a question mark.

%% file: figures/header_prompt.tex
\begin{figure}[t]
    \centering
    \begin{subfigure}[t]{\linewidth}
        \vspace{-2.7cm} 
        \centering
        \includegraphics[width=0.9\linewidth, trim={35.5mm 8cm 4cm 0mm}, clip]{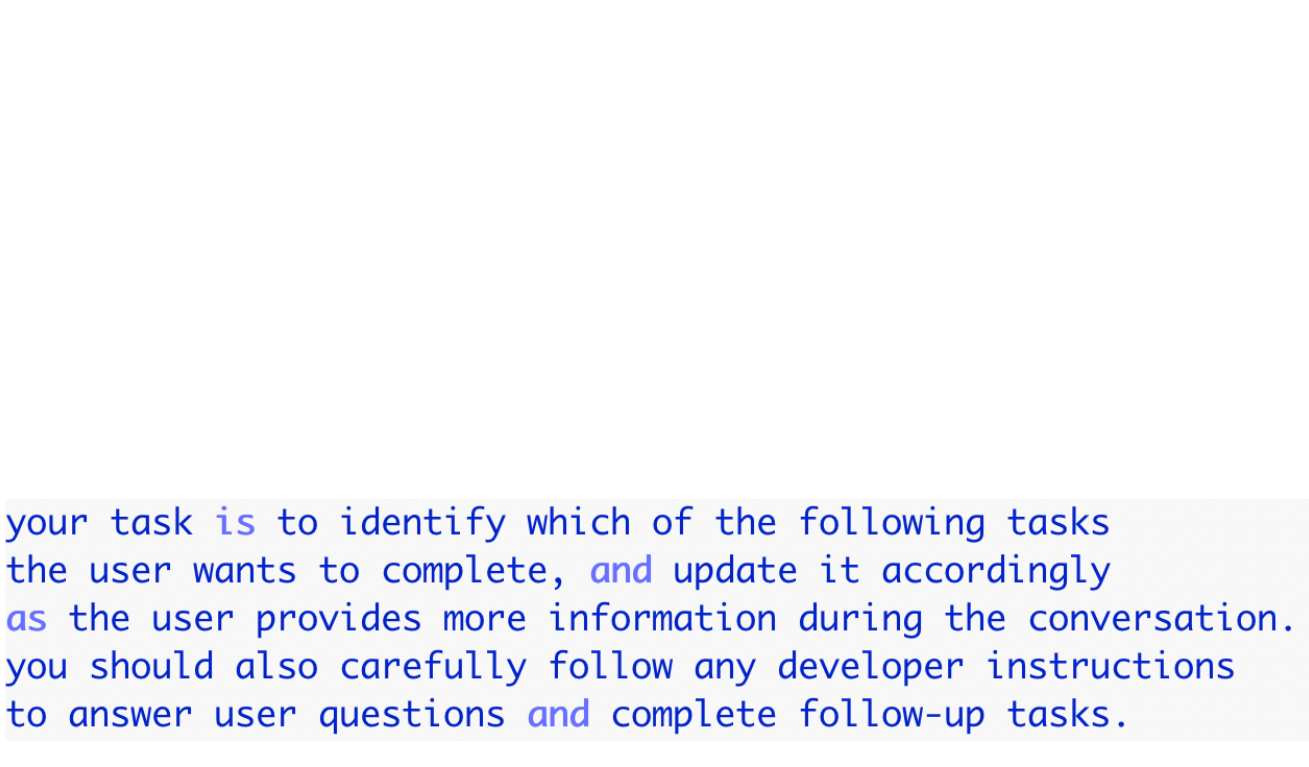}
        \caption{Task instruction}
        \label{fig:task_instruction}
    \end{subfigure}
    \begin{subfigure}[t]{\linewidth}
        \vspace{0.3cm} 

        \centering
        \includegraphics[width=0.9\linewidth, trim={40mm 0cm 4cm 0mm}, clip]{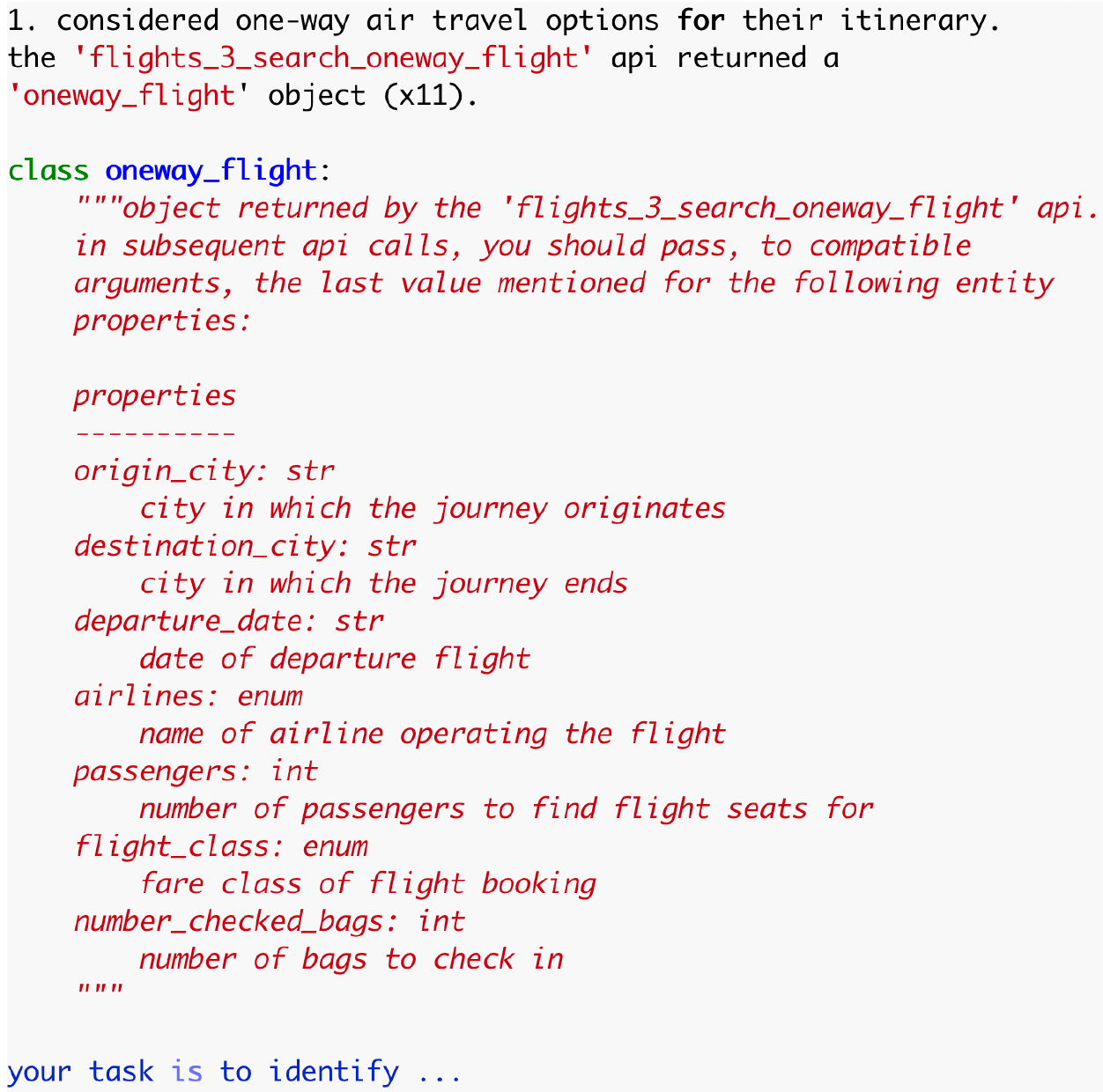}
        \caption{Task stack}
        \label{fig:task_stack}
    \end{subfigure}
    \caption{Action parser header prompt components.}
    \label{fig:action_parser_prompt}
\end{figure}

%% file: figures/context_dependent_instructions.tex
\begin{figure}[t]
    \centering
    \begin{subfigure}[t]{\linewidth}
        \centering
        \includegraphics[width=\linewidth, trim={0mm 0mm 0mm 0mm}, clip]{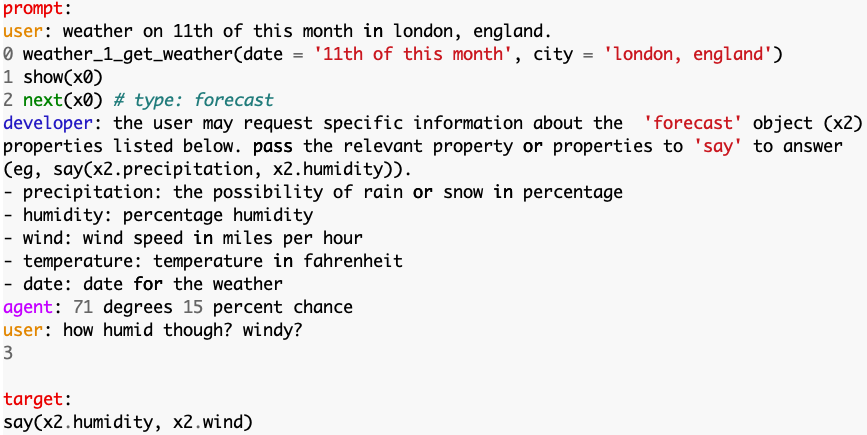}
        \caption{Post-iteration context-dependent instructions.}
        \label{fig:cd_instr_qa}
    \end{subfigure}
    \begin{subfigure}[t]{\linewidth}
        \centering
        \includegraphics[width=\linewidth, trim={17mm 40mm 15mm 40mm}, clip]{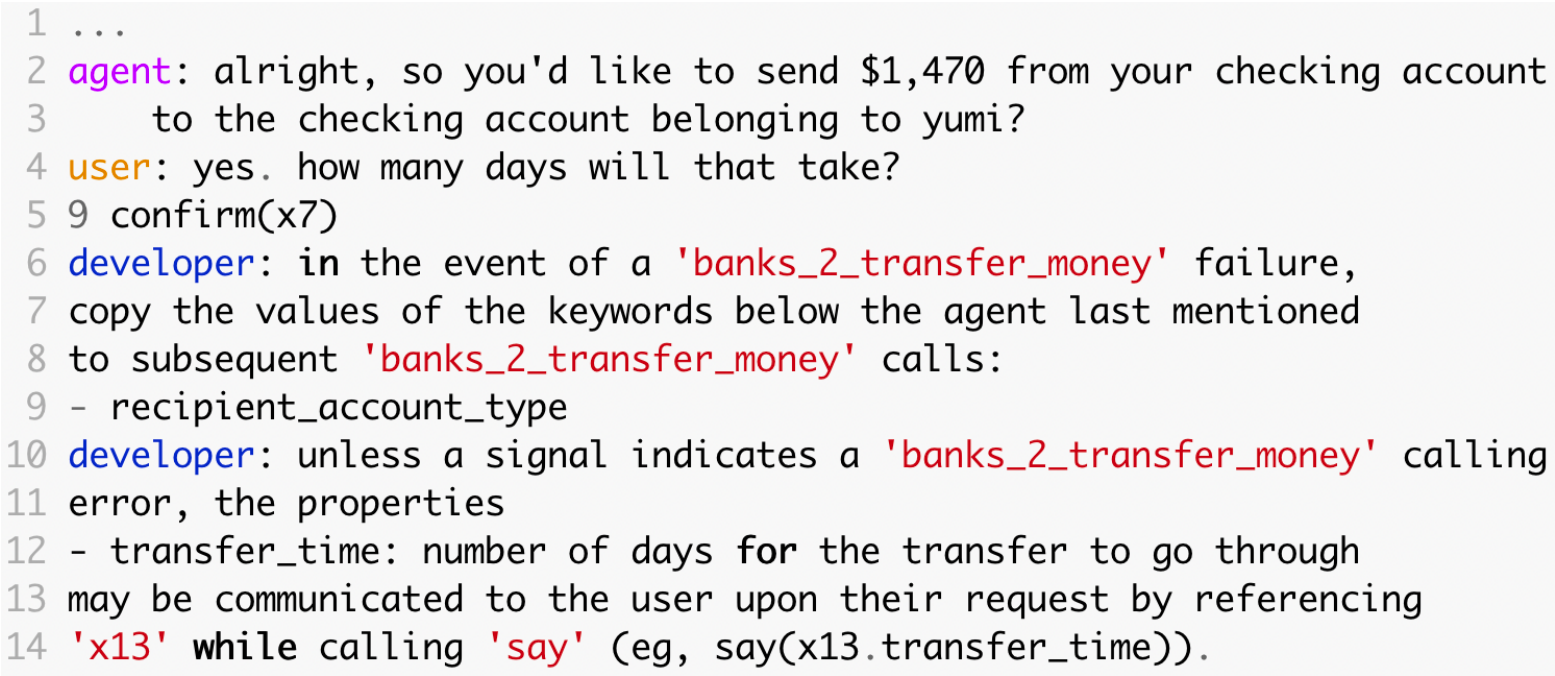}
        \caption{Post-confirmation context-dependent instructions.}
        \label{fig:cd_instr_policy}
    \end{subfigure}
    \caption{Sample context-dependent instructions.}
    \label{fig:context_dependent_instructions}
\end{figure}

%% file: figures/transcript_details.tex
\begin{figure}[t]
\includegraphics[width=0.8\linewidth, trim={10mm 40mm 10mm 43mm}, clip]{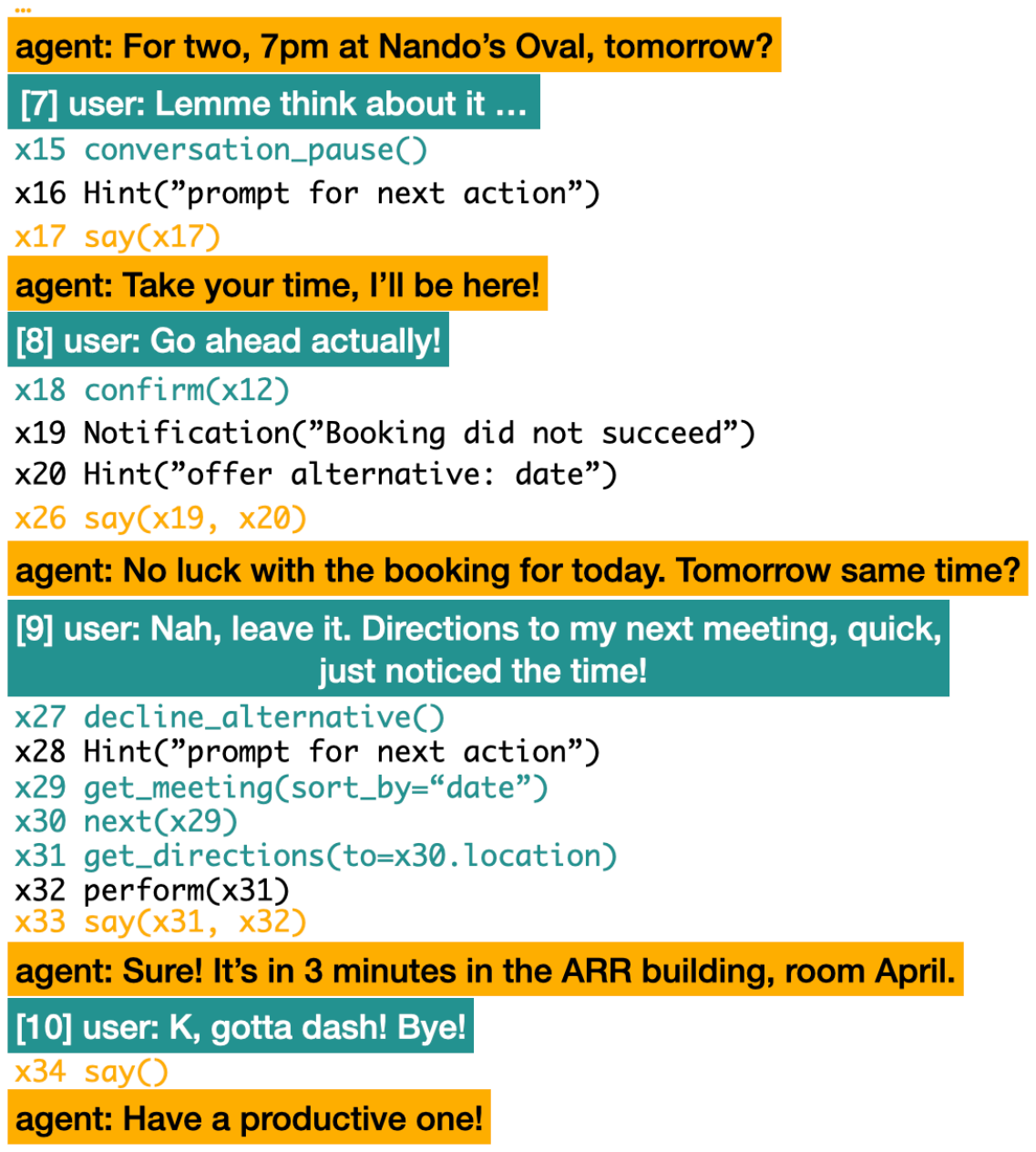}
    \centering
    \caption{An alternative continuation of the conversation in \Cref{fig:transcript} after turn 6. The \texttt{perform} system action an example of how PyTOD parses compositional utterances (turn 9) are depicted. Unlike in \Cref{fig:transcript} where some of the \texttt{say} calls were omitted for clarity, all calls to the NLG module are shown.   } 
    \label{fig:transcript_details}
    \vspace{-0.2cm}
\end{figure}

%% file: figures/ss_templates.tex
\begin{figure}[h]
    \centering

    \begin{subfigure}[b]{0.95\columnwidth}
        \centering
        \includegraphics[width=\linewidth, trim={0mm 55mm 0mm 60mm}, clip]{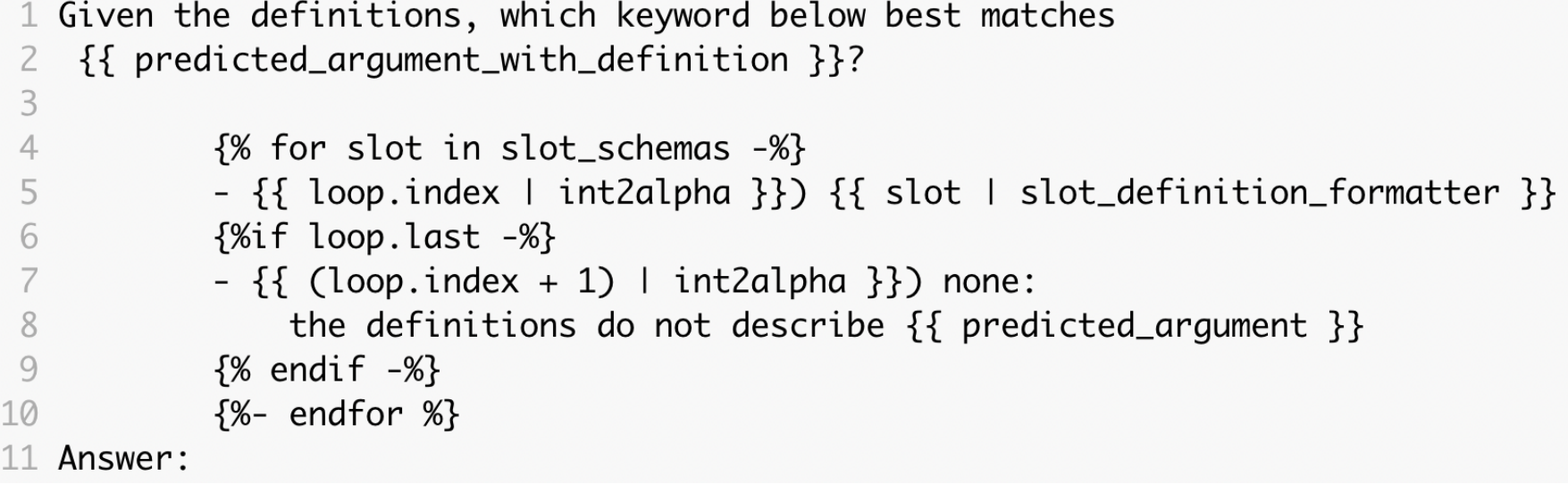}
        \caption{Unknown slot name. Sample prompt in \Cref{fig:unk_slot_prompt}.}
        \label{fig:unk_arg}
    \end{subfigure}


    \begin{subfigure}[b]{0.95\columnwidth}
        \centering
        \includegraphics[width=\linewidth, trim={0mm 30mm 0mm 30mm}, clip]{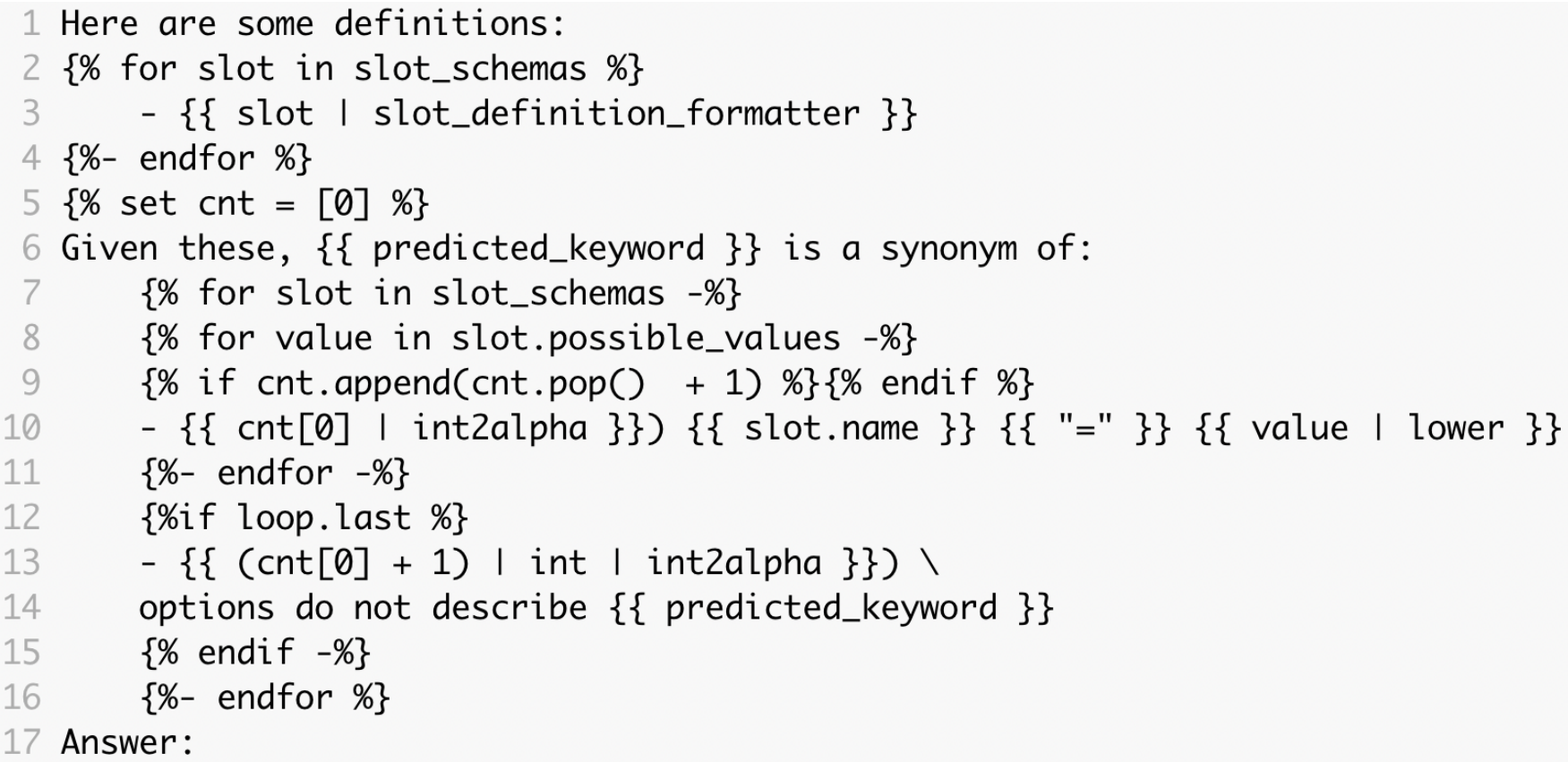}
        \caption{Unknown slot name (closed value). Sample prompt in \Cref{fig:unk_slot_cat_value_prompt}.}
        \label{fig:unk_arg_closed_val}
    \end{subfigure}


    \begin{subfigure}[b]{0.95\columnwidth}
        \centering
        \includegraphics[width=\linewidth, trim={0mm 60mm 0mm 60mm}]{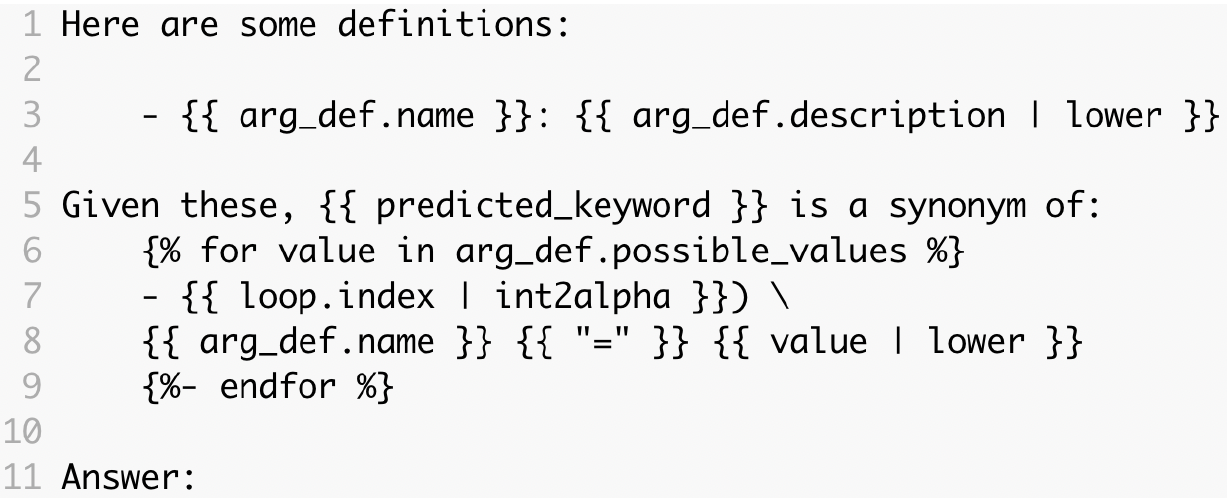}
        \caption{Unknown categorical slot value}
        \label{fig:unk_cat_value}
    \end{subfigure}


    \begin{subfigure}[b]{0.95\columnwidth}
        \centering
        \includegraphics[width=\linewidth, trim={0mm 60mm 0mm 60mm}, clip]{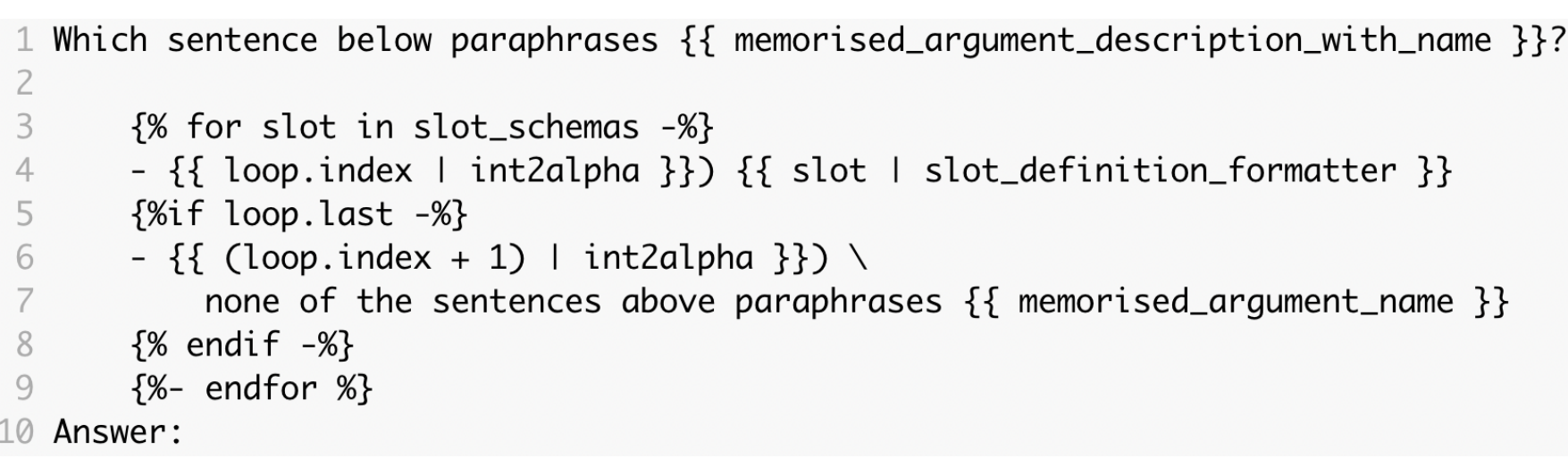}
        \caption{Memorised slot name. Sample prompt in \Cref{fig:memorisation_prompt}.}
        \label{fig:memorizsed_arg}
    \end{subfigure}
    \caption{Schema supervisor prompt templates}
    \label{fig:ss_templates}
\end{figure}

%% file: figures/ps_templates.tex
\begin{figure}[h]
\includegraphics[width=0.95\linewidth, trim={20mm 45mm 20mm 45mm}, clip]{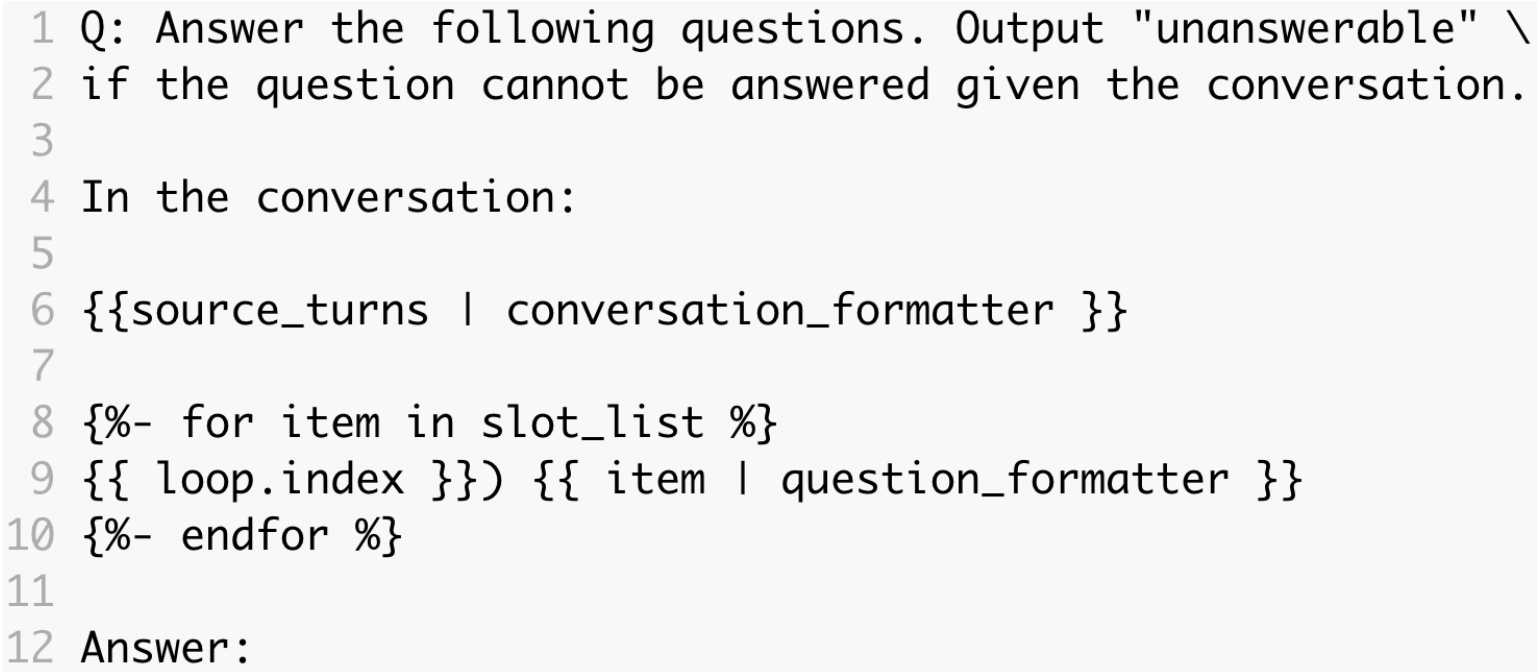}
    \centering
    \caption{Parser supervisor template}
    \label{fig:ps_template}
\end{figure}

%% file: appendices/experimental_setup_v0.2.tex
\section{Experimental Details}
\label{appendix:experiments}
\subsection{Slot values normalisation}
\label{appendix:normalisation}
In SGD, the dialogue state updates when the user either explicitly provides a slot value or accepts a system-proposed value. Traditional TOD agents track the latter  by extracting slot-value pairs from agent utterances. In contrast, as described in rows 3 and 9 of \Cref{tab:tod_syntax_final} (in \S \ref{sec:transcript}), PyTOD updates the state by executing \texttt{select} and \texttt{confirm} commands. These commands read relevant slot values from entities retrieved via database queries (for search-based interactions) or API responses (for transactional interactions). Since SGD system call annotations\footnote{These include call parameters, entity properties, and API responses.} are canonicalised, PyTOD must normalise open-valued parameters extracted from the dialogue history before making API calls and ensure that system-proposed slot values are de-normalised for evaluation.

In practice, normalisation is performed by looking up the surface form of a predicted value in a mapping that links surface forms to their canonical counterparts. This table is easily constructed from SGD semantic annotations, as illustrated in \Cref{fig:sem_frame}. Instead of de-normalising slot values copied from entities or API responses, we equivalently extend the state annotations to include their canonical forms. This ensures that slots tracked via execution are directly comparable to the reference values used by the DSTC8 evaluator.
\input{figures/semantic_frame}

\subsection{PyTOD implementation} 
\label{sec:pytod-impl}
We finetune PyTOD with the hyperparameters in \Cref{tab:pytod_hyperparams} until the development set JGA for \textit{unseen} services is maximised. The learning rate is constant, with no scaling. We follow the same protocol when finetuning \texttt{google/flan-t5-large}, except that we allocate a training budget of just one epoch. All reported results are an average of three runs with different random seeds.
\begin{table}[t]
\centering
\small
\begin{tabular}{lc}
\toprule
\textbf{Hyperparameter} & \textbf{Value} \\
\midrule
Pretrained model & \texttt{google/flan-t5-base} \\
Optimizer & Adafactor \\
Batch size & 32 \\
Learning rate & 0.0001 \\
Warm-up steps & 1500 \\
Number of epochs & 2 \\
Evaluation frequency & 1500 steps \\
\bottomrule
\end{tabular}
\caption{PyTOD training hyperparameters.}
\label{tab:pytod_hyperparams}
\end{table}
\paragraph{Action Parser (AP)} The AP is fine-tuned to learn user action prediction and communicative system action prediction based on a dataset derived from the $16,142$ dialogues in the SGD training set. After filtering examples that exceed the maximum source length of $2048$ tokens, the final training set contains $296,013$ examples. 

\paragraph{Parser Supervisor (PS)} In contrast to the SS, the PS is fine-tuned to handle slot omissions and semantic errors. The fine-tuning dataset for the PS is therefore constructed by creating $34,105$ extractive QA prompts similar to the one in \Cref{fig:ps_prompt} from the SGD training split. Several strategies are employed during data creation.

First, the dialogue history in each prompt is filtered to include only turns relevant to the active service, focusing the model on the immediate context. Second, the order of questions is randomized to prevent the model from learning any positional biases.

Finally, the prompts are augmented with negative examples to train the model to avoid hallucination. With a probability of 0.5, each prompt includes questions for slots whose values are not mentioned in the context. To train the model to make fine-grained distinctions, there is a subsequent $50\%$ chance that one of these unanswerable questions will probe for a slot that is semantically similar to one already present (e.g., asking for a check-out date when the user has only provided a check-in date).

The PS shares its parameters with the AP. This allows PyTOD's core logic to be handled by a single, small pre-trained language model, which requests corrections from an LLM only when validation fails. This design, mediated by the Dialogue Manager, balances performance with the inference costs associated with larger models.

The data pre-processing scripts are included in the linked code release.

\subsection{Replication study}
\label{appendix:training-setup}
 
We finetune D3ST and SDT-Seq using the \texttt{transformers} \cite{transformers} library (v4.35.2). With the exception of FlanT5 \cite{flant5} D3ST, which was trained on two NVIDIA A100 GPUs (80GB), all models were trained on a single NVIDIA RTX 3090 GPU (24GB). 

\subsubsection{D3ST} 

We set all training parameters to match those reported by \citet{d3st} and pre-process the data using their official script\footnote{Available at \href{https://bit.ly/4aKe9KL}{https://bit.ly/4aKe9KL}.}.
As the original work does not specify a model selection metric or evaluation frequency, we evaluate every $5,000$ steps and select the best checkpoint based on overall JGA on the development set. Training is terminated early if accuracy does not improve within $15,000$ steps (approximately $3$ epochs).

Our results differ by an absolute $1.7\%$ from the published JGA for the base model (\Cref{tab:d3st_replication}, rows 1 \& 2). We observe a $0.8\%$ improvement on seen services, but a $2.6\%$ drop on unseen services.
\input{tables/d3st_replication}

To rule out overfitting, we increase the evaluation frequency to $900$ steps and select the model maximising the \textit{unseen services} JGA, stopping the training after $1$ epoch of no improvement. However, this leads to a slight performance drop (\Cref{tab:d3st_replication}, rows 2 \& 3). Hence, finetune Flan-T5 (780M) with the best settings, achieving seen services performance on par with the published results but a $4.6\%$ discrepancy on unseen services (rows 4 \& 5).

\subsubsection{SDT-Seq} 

We set training hyperparameters to the values reported by \citet{sdt}, using the data processing scripts from the official SDT code release\footnote{Available at \href{https://bit.ly/4aKe9KL}{https://bit.ly/4aKe9KL}.}. Since the evaluation frequency and model selection metric are unspecified, we evaluate every $1600$ steps, selecting the model with the highest development set overall JGA. We closely replicate the reported results (\Cref{tab:sdt_seq_replication}).

\input{tables/sdt_replication}

%% file: figures/semantic_frame.tex
\begin{figure}[t]
\includegraphics[scale=0.38, trim={20mm 0mm 20mm 0mm}, clip]{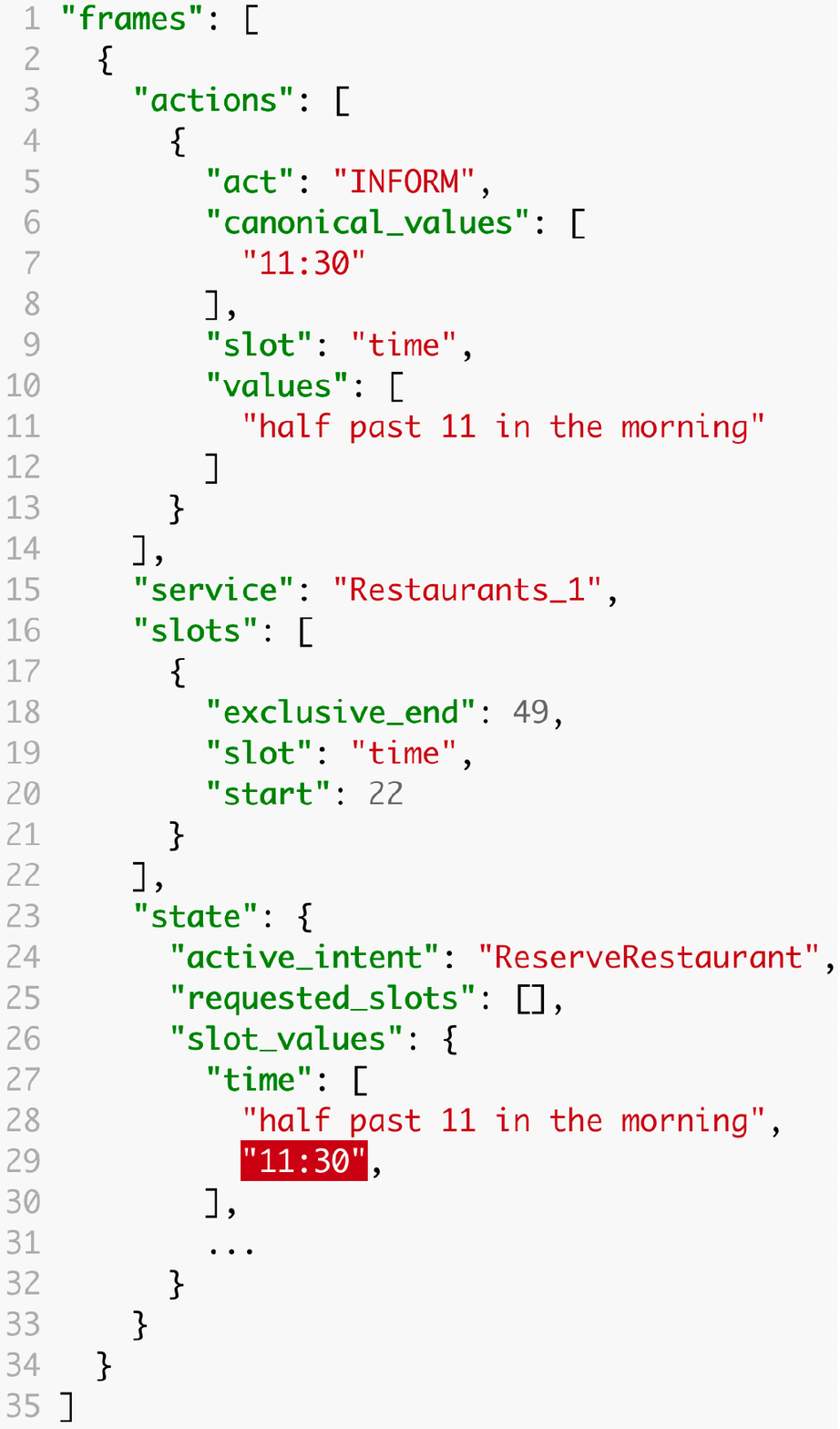}
    \centering
    \caption{Semantic frame for the utterance \textit{I would like it to be half past 11 in the morning.} The action annotations (line 3 - 14) are processed to extend slot-value annotation with the corresponding canonical value (e.g., line 29). The ellipsis in line 31 marks slot-value pairs which were omitted for clarity. }
    \label{fig:sem_frame}
\end{figure}

%% file: tables/d3st_replication.tex
\begin{table}[t]
\centering
\resizebox{\columnwidth}{!}{%
\begin{tabular}{clccccc}
\toprule
\textbf{Size} & \textbf{Model} & \textbf{JGA} & \textbf{JGA Seen} & \textbf{JGA Unseen} & \textbf{Setting} & \textbf{\#} \\ 
\midrule
\multirow{3}{*}{220M} 
& D3ST \cite{d3st} & 72.9 & 92.5 & 66.4 & - & 1 \\
& D3ST (Flan-T5, ours) & 71.2 & 93.2 & 63.8 \deemph{(65.0/64.8/61.7)} & A & 2 \\
& D3ST (Flan-T5, ours) & 70.7 & 92.9 & 63.3  \deemph{(61.9/65.3/62.7)} & B & 3 \\
\midrule
\multirow{2}{*}{780M} 
& D3ST \cite{d3st} & 80.0 & 93.8 & 75.4 & - & 4 \\
& D3ST (Flan-T5, ours) & 76.5 & 93.8 & 70.8 \deemph{(69.9/71.6/70.9)} & A & 5\\ 
\bottomrule
\end{tabular}%
}
\caption{D3ST replication results. Numbers in brackets show the metric values for each experiment run, three-runs averages are shown otherwise.}
\label{tab:d3st_replication}
\end{table}

%% file: tables/sdt_replication.tex
\begin{table}[h]
\centering
\resizebox{\columnwidth}{!}{%
\begin{tabular}{clcccc}
\toprule
\textbf{Size} & \textbf{Model} & \textbf{JGA} & \textbf{JGA Seen} & \textbf{JGA Unseen} & \textbf{\#} \\ 
\midrule
\multirow{2}{*}{220M} 
& SDT-Seq \cite{sdt} & 76.3 & - & - & 1 \\
& SDT-Seq (Flan-T5, ours) & 77.5 & 93.5 & 72.2 & 2 \\
\midrule
\multirow{2}{*}{780M} 
& SDT-Seq \cite{sdt} & 83.3 & - & - & 3 \\ 
& SDT-Seq (Flan-T5, ours) & 82.7 & 94.1 & 78.9 & 4 \\ 

\bottomrule
\end{tabular}%
}
\caption{SDT-Seq replication results.The reported numbers are averaged over five runs, each using a distinct set of demonstrations to construct the fine-tuning prompts.}
\label{tab:sdt_seq_replication}
\end{table}

%% file: appendices/additional_results_v0.1.tex
\section{Extended Analysis and Discussion}
\label{appendix:additional_results}
\subsection{Annotation errors}
\label{appendix:annotation}
\input{tables/annotation_errors_v0.2}
\Cref{tab:mismatched_annotations} presents sample errors identified in our analysis in \S \ref{sec:err-analysis}, showing intent paraphrase errors in \#2\&3. While our DM performs argument-based disambiguation to identify intent, \textit{GetTrainTickets} and \textit{FindTrains},  the two \textit{Trains\_1} intents, share all their arguments. Consequently, misparaphrased intent annotations prevent PyTOD from retrieving train schedules, leading to degraded DST performance. Analysis of $30$ additional dialogues from this domain, we found that  intent confusion caused state errors in $20$ out of $50$ cases.

\Cref{tab:mismatched_annotations} highlights some utterances contain slots mentions without corresponding user action annotations (\#1) while others sometimes fail to paraphrase actions parametrised by boolean or categorical slots (\#4). These issues were difficult to identify with the methods available to \citet{SGD}. During PyTOD development we, however, identified $451$ dialogues across the train, development and test releases\footnote{We include dialogue IDs in our code release.} while experimenting with fault-tolerant execution. This suggests that programmable dialogue systems, while challenging to develop, can serve as valuable tools for dataset quality improvements. 

\subsection{C-JGA breakdown}
\label{appendix:cjga-breakdown}
To support future comparisons, we provide a breakdown of C-JGA metrics for our models and replicated baselines across seen and unseen services.
\input{tables/c_jga_breakdown}

%% file: tables/annotation_errors_v0.2.tex
\newcolumntype{L}[1]{>{\raggedright\arraybackslash}p{#1}} 
\newcolumntype{C}[1]{>{\centering\arraybackslash}p{#1}}   
\newcolumntype{R}[1]{>{\raggedleft\arraybackslash}p{#1}}  

\begin{table}[h]
\centering
\setlength{\tabcolsep}{2pt}

\resizebox{\columnwidth}{!}{%
\begin{tabular}{@{}L{0.2cm} L{3.2cm} L{3.2cm} L{3.7cm} L{1.2cm}@{}}
\toprule
\textbf{\#} 
& \textbf{Utterance} 
& \textbf{Annotation} 
& \textbf{Explanation} 
& \textbf{Service} \\
\midrule

 \multirow{2}{*}{1}
& Today at 2 in the afternoon.
& \multirow{2}{2.8cm}{\textit{pickup\_time='2 in the afternoon'}}
& \multirow{2}{2.8cm}{\textit{start\_date='today'} in utterance}
& \multirow{2}{*}{RC\_3} \\

\midrule
\multirow{3}{*}{2} 
& I'm in the mood for some music and would like to play some songs.
& \multirow{3}{3.2cm}{\textit{intent=LookUpMusic}}
&  \multirow{3}{3.7cm}{The utterance semantics is better represented as \textit{intent=PlayMedia}}.
&  \multirow{3}{*}{MUS\_3} \\

\midrule
\multirow{2}{*}{3} 
& I need a train ticket with a fully refundable feature.
& \multirow{2}{3.2cm}{\textit{intent=FindTrains}\\\textit{class='Flexible'}}
& The utterance semantics is better represented as \textit{intent=GetTrainTickets}.
& \multirow{2}{*}{TR\_1} \\

\midrule
\multirow{4}{*}{4} 
& Okay, what about attractions there. I need Place of Worship, and something with no entry fee.
& \multirow{4}{3.2cm}{\textit{intent=FindAtractions}\\\textit{category="Place of Worship"}\\\textit{free\_entry=True}\\\textit{good\_for\_kids=True}}
& \multirow{4}{3.7cm}{No \textit{good\_for\_kids=True} mention in utterance.}
& \multirow{4}{*}{TRA\_1} \\

\bottomrule
\end{tabular}
}
\caption{Sample annotation errors identified during the error analysis.
\textit{RC\_3=RentalCars\_3}, 
\textit{MUS\_3=Music\_3}, 
\textit{TR\_1=Trains\_1}, 
\textit{TRA\_1=Travel\_1}.}
\label{tab:mismatched_annotations}
\end{table}

%% file: tables/c_jga_breakdown.tex
\begin{table}[h]
\centering
\resizebox{\columnwidth}{!}{%
\begin{tabular}{clcccc}
\toprule
\textbf{Size} & \textbf{Model} & \textbf{C-JGA} & \textbf{C-JGA (Seen)} & \textbf{C-JGA (Unseen)} & \textbf{\#} \\ 
\midrule
\multirow{2}{*}{220M} 
& D3ST (Flan-T5, ours) & 62.2 & 86.0 & 54.3 & 1 \\
& SDT-Seq (Flan-T5, ours) & 68.7 & 86.6 & 62.8 & 2 \\
\midrule
\multirow{2}{*}{780M} 
& D3ST (Flan-T5, ours)  & 66.5 & 87.9 & 61.0 & 3 \\ 
& SDT-Seq (Flan-T5, ours) & 74.2 & 88.0 & 69.6 & 4 \\ 
\midrule
220M & PyTOD (Base) & 72.7 & 87.3 & 67.8 & 5 \\
780M & PyTOD (Large) & 78.4 & 89.1 & 74.9 & 6 \\
\bottomrule
\end{tabular}%
}
\caption{Breakdown of C-JGA from \Cref{tab:jga_comparison}.}
\label{tab:cjga_breakdown}
\end{table}

%% file: appendices/annotated_transcripts.tex
\clearpage
\onecolumn
\section{Annotated Session Transcript Examples}
\label{appendix:annotated-session-transcripts}
\input{tables/ontology_summary}
\twocolumn

%% file: tables/ontology_summary.tex
\renewcommand{\arraystretch}{1.25}

{
\scriptsize 

\begin{longtable}{
  >{\raggedright\arraybackslash}m{0.34\textwidth}  
  >{\raggedright\arraybackslash}m{0.6\textwidth}  
}

\toprule
\textbf{Context} & \textbf{Transcript} \\
\midrule
\endfirsthead

\multicolumn{2}{c}{\textbf{\tablename\ \thetable{} -- continued from previous page}} \\ 
\toprule
\textbf{Context} & \textbf{Transcript} \\
\midrule
\endhead
\multicolumn{2}{r}{\textbf{Continued on next page}} \\ 

\endfoot
\bottomrule
\addlinespace[0.5em]  
\caption{Further session transcript examples, listing the entire set of PyTOD system and user actions along with their semantics. Explanations provide detailed, step-by-step explanations of transcript generation. As the conversation advances, the user and system utterances and the transcript updates of all previous turns are concatenated to form the context for predicting the transcript updates given the current turn (formed of the last system and user utterances).}\label{tab:dialogue}\\
\endlastfoot

\textbf{U\textsubscript{1}:} Please make a payment to Mary. 
& 
\texttt{x1 make\_payment(receiver="Mary")}\newline
\texttt{x2 Hint("Ask user to provide: amount")}\newline
\texttt{x3 Hint("Ask user to provide: private\_transaction")}\newline
\texttt{x4 say(x2)}
\\
\multicolumn{2}{p{\dimexpr\textwidth-2\tabcolsep}}{%
\textbf{Transcript explanation:}\newline
- The AP parses turns where the user introduces a new task to \textit{draft} API calls (\texttt{make\_payment)}.\newline
- The SS is triggered if API call parameters are not in the schema (e.g., \textit{recipient} is generated instead of \textit{receiver}).\newline
- All transcript entries are  bound to a variable (e.g., \texttt{x1}) so that they can be referenced by other instructions. \newline 
- Unless all required arguments are known, API execution returns possible system actions, termed \textit{slot filling hints}. To make a payment, the user must provide a recipient, amount and specify if they wish to make a private transaction or not. No value is known for the last two slots so executing the API recommends these two actions. \newline
- PyTOD executes a recommended action by passing its corresponding variable to the NLG call. Here it chooses to request the payment amount so \texttt{x2} is passed to \texttt{say}. Equivalently, it could have chosen to ask the user if they wish the transaction to be private (in which case \texttt{say(x3)} would have been generated) or requested both slots at once (case in which \texttt{say(x2, x3)} would have been generated).
}\\
\midrule
\textbf{S\textsubscript{2}:} How much?\newline \textbf{U\textsubscript{2}:} Pay her sixty dollars.
& 
\texttt{x5 x1.amount="sixty dollars"}\newline
\texttt{x6 Hint("Ask user to provide: private\_transaction")}\newline
\texttt{x7 say(x6)}
\\
\multicolumn{2}{p{\dimexpr\textwidth-2\tabcolsep}}{%
\textbf{Transcript explanation:}\newline
- The system utterance \textbf{S\textsubscript{2}} is the output of the \texttt{say(x2)} call in the previous turn. We do not consider end-to-end evaluation so in our evaluation we use the ground truth system and user utterances from the corpus. \newline
-  The AP parses the user utterance to an assignment instruction, which is PyTOD's slot filling mechanism. \newline
- Since \texttt{x1} is bound to \texttt{make\_payment}, the assignment updates the properties of the draft command. \newline
- For efficiency, the action recommendations and \texttt{say} are not shown in the prompt for user action parsing so the dialogue context for assignment prediction is \textbf{U\textsubscript{1}}, \texttt{x1}, \textbf{S\textsubscript{2}} and \textbf{U\textsubscript{2}} (along with the AP header (\S \ref{sec:header}), as shown in \Cref{fig:pytod}).\newline
- Given the \texttt{say(x2)} call, the DM anticipates the user will provide the amount in the next turn and uses this information for verification. Missing information would trigger a PS call.\newline 
- After the update, the \texttt{private\_transaction} required argument is still not known, so when the assignment is executed, the API recommends a system action (\texttt{x6}) to find the value of this parameter.
}\\

\midrule
\multicolumn{2}{c}{\textit{New conversation (Restaurant Search \& Booking)}}\\
\midrule

\textbf{U\textsubscript{1}:} I'm after a Chinese in central London.
& 
\texttt{x1 find\_restaurant(area="centre", city="London", cuisine="Chinese")}\newline
\texttt{x2 show(x1)}\newline
\texttt{x3 next(x1)}\newline
\texttt{x4 say(x3)}
\\
\multicolumn{2}{p{\dimexpr\textwidth-2\tabcolsep}}{%
\textbf{Transcript explanation:}\newline
- When all the required arguments to the \texttt{find\_restaurant} search query are known, execution returns a list of restaurants matching the user request, which is bound to \texttt{x1}.\newline
- The dialogue manager signals that search returned some results by inserting \texttt{show(x1)} in the transcript. The only positional argument of \texttt{show} is a reference to the API call that returned the results.\newline
- Given \textbf{U\textsubscript{1}}, and the transcript lines \texttt{x1} and \texttt{x2}, PyTOD generates the instructions bound to \texttt{x3} and \texttt{x4}.\newline 
- \texttt{next(x1)} fetches the top returned entity (a \texttt{Restaurant} object) and binds it to a variable (\texttt{x3}).\newline
- Entities can be passed to NLG calls as positional arguments (\texttt{x4}). 
}\\
\midrule
\textbf{S\textsubscript{2}:} There are many options, the first is Golden Wok, 500 ft away.\newline \textbf{U\textsubscript{2}:} What else?
& 
\texttt{x5 next(x1)}\newline
\texttt{x6 say(x5)}
\\
\multicolumn{2}{p{\dimexpr\textwidth-2\tabcolsep}}{%
\textbf{Transcript explanation:}\newline
- When called with an entity variable, the NLG generates an utterance that provides the user with information relevant in context.\newline
- When then user requests an alternative, the results list is iterated using a \texttt{next} instruction (\texttt{x5}) and the result is communicated to the user (\texttt{x6}).
}\\
\midrule
\textbf{S\textsubscript{3}:} The Burnt Kidney?\newline \textbf{U\textsubscript{3}:} The first one sounds like a winner.
& 
\texttt{x7 select(x3, from\_results=x1)}\newline
\texttt{x8 suggest(task="book\_restaurant")}\newline
\texttt{x9 Hint("Prompt user for next task")}\newline
\texttt{x10 say(x8)}
\\
\multicolumn{2}{p{\dimexpr\textwidth-2\tabcolsep}}{%
\textbf{Transcript explanation:}\newline
- The AP generates the \texttt{select} statement 
when the user selects one of the alternatives proposed by
the agent\newline
- The \texttt{select} positional argument is bound to a search result fetched with \texttt{next}.\newline
- \texttt{from\_results} is a reference to the query that returned the list of results.\newline
- \texttt{select} execution returns two system action recommendations: a \textit{suggestion} (\texttt{x8}) and a \textit{task prompting hint} (\texttt{x9}).\newline
-  \texttt{x8} binds a \texttt{book\_restaurant} draft command updated with relevant details from history (e.g., \texttt{restaurant\_name, location}).\newline
- The agent chooses to pass the suggestion to the NLG module (\texttt{x10}).
}\\
\midrule
\textbf{S\textsubscript{5}:} Book it now?\newline \textbf{U\textsubscript{5}:} Not now. Tell me my checking balance.
& 
\texttt{x11 suspend(x6)}\newline
\texttt{x12 check\_balance(account="checking")}\newline
\texttt{x13 show(x12)}\newline
\texttt{x14 next(x12)}\newline
\texttt{x15 say(x14)}
\\
\multicolumn{2}{p{\dimexpr\textwidth-2\tabcolsep}}{%
\textbf{Transcript explanation:}\newline
- The AP generates a \texttt{suspend} instruction  when the user declines a suggestion.\newline
- \texttt{suspend} takes variables bound to a \texttt{suggest} statement as input.\newline
- In this case, the AP generates the \texttt{suspend} instruction and \texttt{check\_balance} API call at once.\newline
- \texttt{check\_balance} only requires the \texttt{account} argument, so executing it returns a list of accounts (\texttt{x12}), which are signalled to the agent (\texttt{x13}).\newline 
- Like before, PyTOD generates the instructions \texttt{x14} and \texttt{x15} at once.\newline
- The top result, an \texttt{Account} entity,  is fetched from results (\texttt{x14}).\newline 
- The agent passes the \texttt{Account} entity to the NLG module to provide the user with the relevant information (\texttt{x15})
}\\
\midrule
\textbf{S\textsubscript{6}:} There are \$10,535.55.\newline \textbf{U\textsubscript{6}:} Okay, let’s book a table there! Next Monday, at 7.
& 
\texttt{x16 resume(x11)}\newline
\texttt{x17 x16.date="Next Monday"; x16.time="7"}\newline
\texttt{x18 Hint("Confirm: restaurant\_name, date, time")}\newline
\texttt{x19 say(x18)}
\\
\multicolumn{2}{p{\dimexpr\textwidth-2\tabcolsep}}{%
\textbf{Transcript explanation:}\newline
- The AP parses the last user utterance to two actions, bound to \texttt{x16} and \texttt{x17}.\newline
- When the user refers to a task they previously suspended, a \texttt{resume} statement whose only positional argument is a variable bound to a \texttt{suspend} statement is generated.\newline
- Executing a \texttt{resume} statement returns a draft API call updated with relevant parameters from the dialogue history: a \texttt{book\_restaurant} instance with known \texttt{restaurant\_name} and location in this case (which had been bound to \texttt{x11} when the task was suspended). \newline
- Resumed tasks may be updated through assignments (\texttt{x17}).\newline
- When all the arguments for a transaction are known, draft API call execution returns a  \textit{confirmation hint} (\texttt{x18}).\newline
- The agent requests the user to confirm by passing the confirmation hints to the NLG module (\texttt{x19).}
}\\
\midrule
\textbf{S\textsubscript{7}:} Please confirm your Golden Wok booking, 7th April at 7 pm.\newline \textbf{U\textsubscript{7}:} You got it!
& 
\texttt{x20 confirm(x16)}\newline
\texttt{x21 Notification("Booking did not succeed")}\newline
\texttt{x22 Hint("Alternative: time=9:15pm")}\newline
\texttt{x23 say(x21, x22)}
\\
\multicolumn{2}{p{\dimexpr\textwidth-2\tabcolsep}}{%
\textbf{Transcript explanation:}\newline
- When the user confirms a transaction, the AP generates a \texttt{confirm} statement whose positional argument is a 
variable bound to the draft API. \newline
- \texttt{confirm} calls execute the API. Task completion failures are indicated by \texttt{Notification} messages (\texttt{x21}).\newline
- If transactions can be completed given an alternative set of constraints, an \textit{alternative hint} system action indicating possible alternatives is returned by the command (\texttt{x22}).\newline
- Task status and alternatives are communicated to the user (\texttt{x23}).
}\\
\midrule
\textbf{S\textsubscript{8}:} Only 9:15 is available. Happy with that?\newline \textbf{U\textsubscript{8}:} I’m afraid not.
& 
\texttt{x24 decline\_alternative()}\newline
\texttt{x25 Hint("Prompt user for next task")}\newline
\texttt{x26 say(x25)}
\\
\multicolumn{2}{p{\dimexpr\textwidth-2\tabcolsep}}{%
\textbf{Transcript explanation:}\newline
 - The \texttt{decline\_alternative} instruction is generated by the AP to mark negation.\newline
 - Executing \texttt{decline\_alternative} returns task prompting hints (\texttt{x25}) which are communicated to the user (\texttt{x26}).
}\\
\midrule
\textbf{S\textsubscript{9}:} Anything else I can do for you?\newline \textbf{U\textsubscript{9}:} Let me think...
& 
\texttt{x27 conversation\_pause()}\newline
\texttt{x28 Hint("Prompt user for next task")}\newline
\texttt{x29 say(x28)}
\\
\multicolumn{2}{p{\dimexpr\textwidth-2\tabcolsep}}{%
\textbf{Transcript explanation:}\newline
- The AP generates \texttt{conversation\_pause} instructions to mark out-of-domain utterances or situations when the user does not take an action. 
}\\

\midrule
\multicolumn{2}{c}{\textit{New conversation (Search \& Play a Movie)}}\\
\midrule

\textbf{U\textsubscript{1}:} I want to watch a Disney movie.
& 
\texttt{x1 find\_movie(producer="Disney")}\newline
\texttt{x2 slice(x1)}\newline
\texttt{x3 say(x2)}
\\
\multicolumn{2}{p{\dimexpr\textwidth-2\tabcolsep}}{%
\textbf{Transcript explanation:}\newline
- \texttt{slice} is analogous to \texttt{show} (see above), but binds a small number of entities the agent should preset to the user.\newline
- when called with a slice reference, \texttt{say} produces an utterance where multiple entities are mentioned.
}\\
\midrule
\textbf{S\textsubscript{2}:} Let’s see … Snow White, Frozen 1, or Frozen 2?\newline \textbf{U\textsubscript{2}:} Good taste, let’s go with Frozen 2.
& 
\texttt{x4 select(movie="Frozen 2", from\_results=x1)}\newline
\texttt{x5 suggest(task="play\_movie")}\newline
\texttt{x6 Hint("Prompt user for next task")}\newline
\texttt{x7 say(x5)}
\\
\multicolumn{2}{p{\dimexpr\textwidth-2\tabcolsep}}{%
\textbf{Transcript explanation:}\newline
- \texttt{select} can take an entity name (e.g., \texttt{movie}) as a keyword argument instead of a positional argument when multiple entities are mentioned in a single utterance.
}\\
\midrule
\textbf{S\textsubscript{3}:} Play it?\newline \textbf{U\textsubscript{3}:} On the living room TV!
& 
\texttt{x8 play\_movie(movie="Frozen 2", device="TV", area="living room")}\newline
\texttt{x9 Hint("Confirm: area, living room")}\newline
\texttt{x10 say(x9)}
\\
\midrule
\textbf{S\textsubscript{4}:} Confirm Frozen 2, in the living room?\newline \textbf{U\textsubscript{4}:} Yes! When did it come out?
& 
\texttt{x11 confirm(x8)}\newline
\texttt{x12 perform(x8)}\newline
\texttt{x13 say(x12, x8.release\_year)}
\\
\multicolumn{2}{p{\dimexpr\textwidth-2\tabcolsep}}{%
\textbf{Transcript explanation:}\newline
- \texttt{perform} takes as a sole argument a variable bound to a transactional intent, and is inserted in the transcript by the DM to mark successful transaction completion.\newline
- An entity (e.g., \texttt{SelectedMovie}) is bound to the transaction variable (\texttt{x8}) upon successful execution.\newline
- Passing variables bound to perform (\texttt{x12}) to the NLG enables the agent to communicate task success to the user.\newline
- Information requested by the user is communicated by passing the entity to the NLG call and specifying the attribute that should be communicated (\texttt{x13}). This is possible since the entity definitions are displayed in the prompt after confirmations, as shown in \Cref{fig:cd_instr_policy}.
}\\
\midrule
\textbf{S\textsubscript{5}:} Playing your movie, released in 2019.\newline \textbf{U\textsubscript{5}:} Thanks, have a good one!
& 
\texttt{x14 say()}
\\
\multicolumn{2}{p{\dimexpr\textwidth-2\tabcolsep}}{%
\textbf{Transcript explanation:}\newline
- The AP generates \texttt{say} with no arguments to mark conversation end.
}\\
\end{longtable}
}
\twocolumn